\documentclass[sigconf]{acmart}
\usepackage{amsmath}
  
\usepackage{amssymb}
\usepackage{enumitem}
\AtBeginDocument{%
  }

\setcopyright{acmlicensed}

\begin{document}

\title{Plan Then Retrieve: Reinforcement Learning-Guided Complex Reasoning over Knowledge Graphs}

\author{Yanlin Song}
\authornote{Equal contribution.}
\affiliation{%
  \department{School of Computer Science}
  \institution{Wuhan University}
  \city{Wuhan}
  \country{China}
}
\email{songyanlin@whu.edu.cn}

\author{Ben Liu}
\authornotemark[1]
\affiliation{%
  \institution{Ant Group}
  \city{Hangzhou}
  \country{China}
}
\email{keli.lb@antgroup.com}

\author{Víctor Gutiérrez-Basulto}
\affiliation{%
  \department{School of Computer Science and Informatics}
  \institution{Cardiff University}
  \city{Cardiff}
  \country{UK}
}
\email{GutierrezBasultoV@cardiff.ac.uk}

\author{Zhiwei Hu}
\affiliation{%
  \department{College of Information Science and Engineering}
  \institution{Shanxi Agricultural University}
  \city{Taiyuan}
  \country{China}
}
\email{zhiweihu@whu.edu.cn}

\author{Qianqian Xie}
\authornotemark[2]
\affiliation{%
  \department{School of Artificial Intelligence}
  \institution{Wuhan University}
  \city{Wuhan}
  \country{China}
}
\affiliation{%
  \department{Center for Language and Information Research}
  \institution{Wuhan University}
  \city{Wuhan}
  \country{China}
}
\email{xieq@whu.edu.cn}

\author{Min Peng}
\authornote{Corresponding Author.}
\affiliation{%
  \department{School of Artificial Intelligence}
  \institution{Wuhan University}
  \city{Wuhan}
  \country{China}
}
\affiliation{%
  \department{Center for Language and Information Research}
  \institution{Wuhan University}
  \city{Wuhan}
  \country{China}
}
\email{pengm@whu.edu.cn}

\author{Sophia Ananiadou}
\affiliation{%
  \department{School of Computer Science}
  \institution{University of Manchester}
  \city{Manchester}
  \country{UK}
}
\email{sophia.ananiadou@manchester.ac.uk}

\author{Jeff Z. Pan}
\affiliation{%
  \department{ILCC, School of Informatics}
  \institution{University of Edinburgh}
  \city{Edinburgh}
  \country{UK}
}
\email{j.z.pan@ed.ac.uk}

\newcommand{\todo}[1]{\textcolor{red}{[todo: #1]}}

\newcommand{\vic}[1]{\textcolor{black}{#1}}
\newcommand{\zhiwei}[1]{\textcolor{black}{#1}}
\newcommand{\nb}[1]{\textcolor{red}{$\blacktriangleright$}\footnote{\textcolor{blue}{#1}}}

\newcommand{\paperComment}[3]{{\color{#2}\par\noindent\fbox{\parbox{0.97\linewidth}{{\sf #1:} #3}}\newline}}
 \newcommand{\victor}[1]{\paperComment{Victor}{purple!50!black}{#1}}

\renewcommand{\shortauthors}{Yanlin Song et al.}

\begin{abstract}
Knowledge Graph Question Answering (KGQA) aims to answer natural language questions by reasoning over structured knowledge graphs (KGs). While large language models (LLMs) have advanced KGQA through their strong reasoning capabilities, existing methods continue to struggle to fully exploit both the rich knowledge encoded in KGs and the reasoning capabilities of LLMs, particularly in complex scenarios. 
They often assume complete KG coverage and lack mechanisms to judge when external information is needed, and their reasoning remains locally myopic, failing to maintain coherent multi-step planning, leading to reasoning failures even when relevant knowledge exists.
We propose \textbf{Graph-RFT}, a novel two-stage reinforcement fine-tuning KGQA framework with a “plan–KGsearch–and–Websearch–during–think” paradigm, that enables LLMs to perform autonomous planning and adaptive retrieval scheduling across KG and web sources under incomplete knowledge conditions. 
Graph-RFT introduces a chain-of-thought (CoT) fine-tuning method with a customized plan–retrieval dataset activates structured reasoning and resolves the GRPO cold-start problem.
It then introduces a novel plan–retrieval guided reinforcement learning process integrates explicit planning and retrieval actions with a multi-reward design, enabling coverage-aware retrieval scheduling.
It employs a Cartesian-inspired planning module to decompose complex questions into ordered sub-questions, and logical expression to guide tool invocation for globally consistent multi-step reasoning.
This reasoning–retrieval process is optimized with a multi-reward combining outcome and retrieval-specific signals, enabling the model to learn when and how to combine KG and web retrieval effectively.
Experiments on multiple KGQA benchmarks demonstrate that Graph-RFT achieves superior performance over strong baselines, even with smaller LLM backbones, and substantially improves complex question decomposition, factual coverage, and tool coordination.

\end{abstract}




\maketitle

\section{Introduction}

Knowledge Graph Question Answering (KGQA) aims to answer natural language questions by reasoning over structured knowledge graphs (KGs) that store explicit, factual knowledge in the form of triples. KGQA plays a central role in web-based intelligent systems such as search, recommendation, and social platforms~\cite{khorashadizadeh2024research, zhao2024breaking, zeng2024large, xie2023pixiu, xie2024finben, xie2025medical, wang2023pre}, where accurate and explainable reasoning over structured data is crucial.
Recent advances in large language models (LLMs) have transformed natural language understanding, reasoning, and generation~\cite{hendrycks2020measuring, clark2018think, guo2025deepseek, shi2025search, liu2025visual}. Leveraging their strong generalization and reasoning capabilities, recent KGQA methods increasingly rely on LLMs, integrating knowledge graphs as structured, verifiable knowledge sources to enhance factual precision and interpretability~\cite{sun2023think, jiang2023structgpt, tan2025paths, ma2025debate, liu2025symagent, xue2024decompose}.

\begin{figure}[t]
    \centering
    \includegraphics[width=0.95\linewidth]{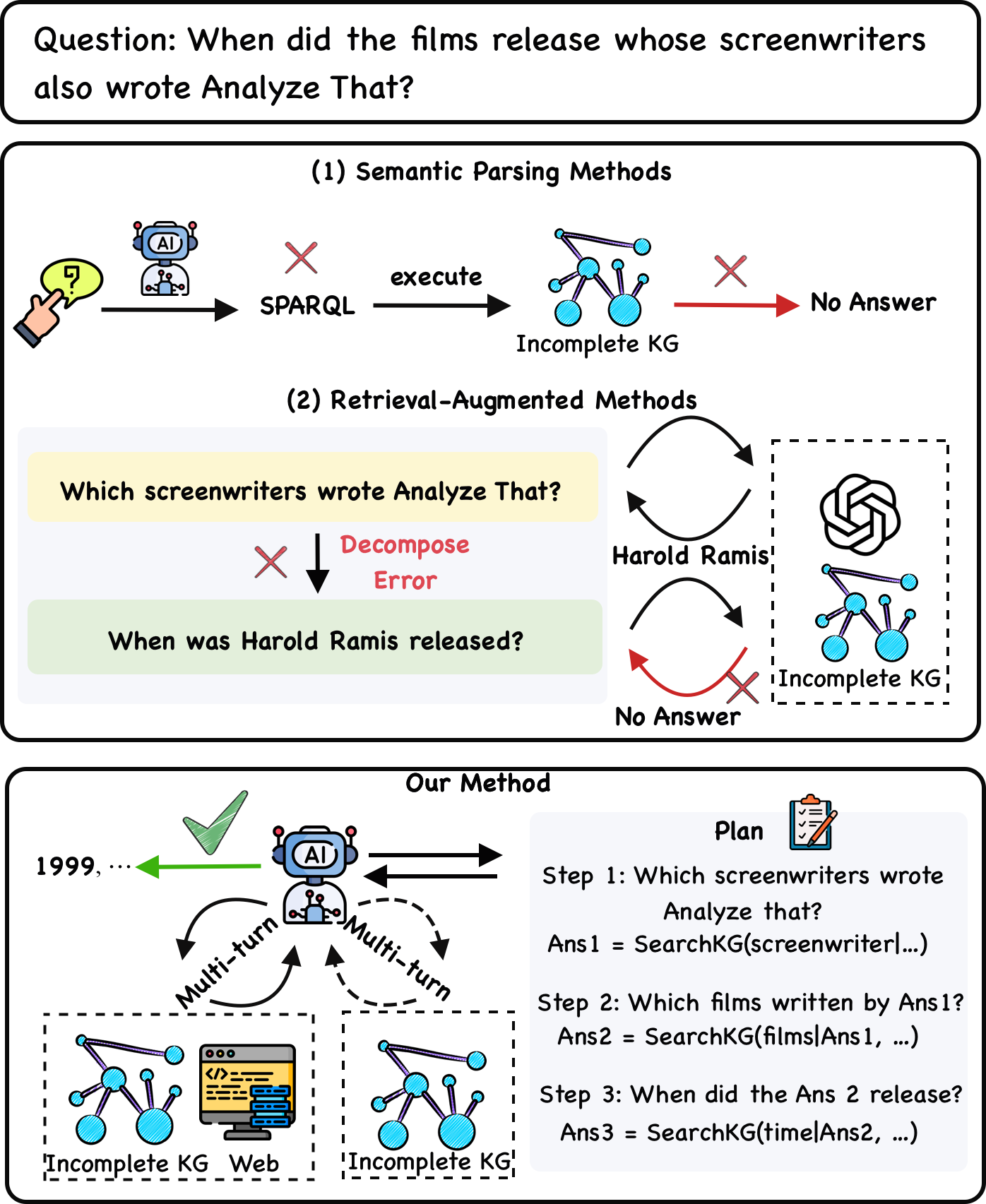}
    \caption{The comparison of Semantic Parsing, Retrieval-Augmented Methods, and Graph-RFT. }
    \label{fig:intro}
\end{figure}

Existing LLM-based KGQA approaches fall into two paradigms: \textbf{semantic parsing (SP)}, which translates natural questions into executable logical queries (e.g., SPARQL) and runs them over a KG \cite{li2023few, luo2023chatkbqa, xu2024generate}, and \textbf{retrieval-augmented generation (RAG)}, which retrieves relevant KG triples to condition an LLM’s generation \cite{tan2025paths, xu2024generate}.
Despite substantial progress, both paradigms struggle in complex settings for two main reasons.
First, they lack a reliable mechanism to judge whether a KG contains the facts necessary to answer a question. SP and RAG methods commonly assume that all required triples are present in the KG, a strong and unrealistic assumption for manually curated graphs. When coverage is incomplete, SP fails to produce correct executable queries, and RAG offers no built-in way to refine retrieval or to decide when to consult external (unstructured) sources.
Second, these methods often lack the ability to perform coherent multi-step planning and adaptive retrieval scheduling. For example in Figure \ref{fig:intro}(2), an LLM may correctly decompose a question into ``Which screenwriters wrote Analyze That?'' and retrieve ``Harold Ramis,'' but then fail to continue the planned decomposition and instead reformulates the follow-up as ``When was Harold Ramis released?'', confusing the person with a film title. This failure stems from constrained local decision-making and an absence of coherent multi-step planning, which leads to retrieval and reasoning errors even when the KG contains the necessary facts. 
The comparison for previous work and Graph-RFT is in Table \ref{intro:compare}.

\begin{sloppypar}

To address these challenges, we propose Graph-RFT, a novel two-stage reinforcement fine-tuning framework designed to enhance an LLM’s capabilities in autonomous planning and adaptive retrieval scheduling across multiple tools (specifically, KG search and web search) under conditions of incomplete KGs.
In the first stage (Section~\ref{data_collection}), we introduce a chain-of-thought (CoT)~\cite{wei2022chain} fine-tuning method that uses a customized plan–retrieval dataset to explicitly train the model to decompose questions, formulate plans, and select retrieval tools. This stage not only activates the model’s reasoning and planning capabilities but also resolves the GRPO \cite{guo2025deepseek} cold-start problem, enabling stable and meaningful reinforcement rollouts. 

In the second stage (Section~\ref{second_stage}), we propose a plan–retrieval guided reinforcement learning method that integrates explicit planning steps and retrieval actions into the model’s reasoning loop, and learns a coverage-aware retrieval policy that dynamically coordinates knowledge graph and web search under incomplete KG conditions.
It uses a structured template
$\langle$\textit{plan}$\rangle\langle$\textit{relation\_search}$\rangle\langle$\textit{neighbor\_search}$\rangle$$\langle$\textit{web\_search}$\rangle$, to explicitly let the model to alternate between symbolic planning and retrieval actions. 
The planning module draws inspiration from Cartesian principles~\cite{descartes1901discourse}, breaking complex questions into logically ordered sub-questions. We further employ logical expressions to represent dependencies among sub-questions and guide the invocation of KG retrieval tools, ensuring globally coherent multi-step reasoning and controllable retrieval flow. For each planned execution step, Graph-RFT invokes both relation-search and neighbor-search tools to perform KG retrieval; when the KG lacks sufficient coverage, the model automatically triggers web search to acquire supplementary unstructured evidence.
To optimize the reasoning–retrieval process, we apply a GRPO-based reinforcement learning with a novel multi-reward design: an outcome reward measuring factual correctness and a retrieval-specific reward that evaluates retrieval coverage, precision, and timing.
This formulation encourages the model to learn when and how to combine KG and web retrieval, overcoming the limitations of prior SP and RAG methods that assume complete KGs and the absence of coherent multi-step planning.

\begin{table}[htbp]
  \centering
  \caption{Comparison for main characteristics of previous SP, RAG methods with our Graph-RFT.}
  \label{intro:compare}
  \resizebox{\linewidth}{!}{
    \begin{tabular}{c|cc|cccc}
      \toprule
      \textbf{Class} & \textbf{Algorithm} & \textbf{Paradigm} & \textbf{Incomplete KG} & \textbf{Global Planning} & \textbf{Decompose Problems} & \textbf{KG Interaction} \\
      \midrule
      SP & RGR-KBQA \cite{feng2025rgr} & SFT & $\times$ & $\times$ & $\times$ & $\times$ \\
      SP & Rule-KBQA \cite{zhang2025rule} & SFT & $\times$ & $\times$ & $\times$ & $\times$ \\
      \midrule
      RAG & PoG \cite{tan2025paths} & ICL & $\times$ & $\times$ & $\checkmark$ & $\checkmark$ \\
      RAG & DoG \cite{ma2025debate} & ICL & $\times$ & $\times$ & $\checkmark$ & $\checkmark$ \\
      RAG & KG-Agent \cite{jiang2024kg} & SFT & $\times$ & $\times$ & $\checkmark$ & $\checkmark$ \\
      \midrule
      Our Method & Graph-RFT & SFT \& RL & $\checkmark$ & $\checkmark$ & $\checkmark$ & $\checkmark$ \\
      \bottomrule
    \end{tabular}
  }
\end{table}

Our main contributions are as follows:
\end{sloppypar}
\begin{itemize}[itemsep=0.5ex, leftmargin=3mm]
\item  We introduce Graph-RFT, a novel two-stage reinforcement fine-tuning framework for complex reasoning over incomplete KGs. It introduces the novel CoT-based fine-tuning method to activate planning and reasoning capabilities while resolving the GRPO cold-start issue, and a plan–retrieval guided reinforcement learning method to enable coherent multi-step reasoning and adaptive retrieval scheduling across KG and web sources under incomplete KG conditions.

\item  We propose a plan–retrieval guided reinforcement learning method with a multi-reward design that integrates graph-retrieval, web-retrieval, and penalty signals. This fine-grained feedback enables the model to learn when and how to combine KG and web retrieval for adaptive, coverage-aware reasoning under incomplete KGs.
\item  Experimental results across several widely used complex reasoning datasets confirm the superior performance of Graph-RFT against strong baselines, even when utilizing weaker LLM backbones (i.e., the 7B series). Furthermore, comprehensive empirical analyses validate Graph-RFT's effectiveness across multiple critical aspects, including complex question decomposition planning, the identification of missing factual triples, and the appropriate invocation of various tools.
\end{itemize}
\section{Related Work}
\vic{Existing methods for improving LLM-based KGQA can be categorized into two main types: Semantic Parsing (SP) methods and Retrieval Augmented (RA) methods.} 

\smallskip \noindent
\vic{\textbf{SP Methods} convert natural language questions into 
structural queries (e.g., SPARQL) 
that can be executed on KGs to 
retrieve answers. Early research efforts \cite{hu2017answering, xiong2017explicit, luo2025chatrule} in semantic parsing often involved generating query graphs, resembling subgraphs of the KG and allow for direct mapping to logical forms. 
Lan \cite{lan2020query} incorporates constraints into the query graph, which effectively narrows the search space and facilitates more flexible query graph generation. 
Subsequent works~\cite{atif2023beamqa, schick2023toolformer, raffel2020exploring, wei2022chain, jiang2022unikgqa, lan2020query} utilize sequence-to-sequence models for multi-hop path and SPARQL-expression generation,  enhancing the semantic parsing process. Recently, with the emergence of LLMs, various methods unify KGs and LLMs for KGQA.
ChatKBQA~\cite{luo2023chatkbqa} generates question-aligned logical forms by treating LLMs as semantic parsers and retrieves entities and relations from KGs. RGR-KBQA~\cite{feng2025rgr} retrieves factual knowledge from KGs to enhance the semantic understanding capabilities of LLMs and finetune them to generate the logical form. 
Rule-KBQA~\cite{zhang2025rule} guides logical form generation via a two-phase process that inducts a rule library and deducts forms via a rule-guided agent. However, these methods depend heavily on the assumption that KGs are complete, which means no answers can be obtained if information is missing.}

\smallskip\noindent 
\textbf{RA Methods} retrieve relevant 
\vic{KG triple based on input questions, then use these triples to guide LLMs to generate accurate answers, emphasizing the dynamic interaction between KG retrieval and LLM reasoning. 
ToG~\cite{sun2023think} employs an explore-and-exploit approach, which can discover the most promising reasoning paths and return the most likely reasoning results. GoG~\cite{xu2024generate}  proposes a think-search-generate paradigm, alleviating the incompleteness issue of KGs by using LLMs' internal knowledge.} 
\vic{PoG~\cite{tan2025paths} leverages knowledge reasoning paths as retrieval-augmented inputs for LLMs, thereby alleviating issues related to insufficient domain knowledge and unfaithful reasoning chains.
DoG~\cite{ma2025debate} introduces a subgraph-focused mechanism that allows LLMs to attempt intermediate answers after each reasoning step, effectively mitigating the drawbacks of lengthy reasoning trajectories.
However, most of these approaches depend on powerful closed-source LLM APIs (e.g., GPT-4~\cite{achiam2023gpt}), which results in substantial performance degradation when weaker open-source models are used as backbones. 
KG-Agent~\cite{jiang2024kg} develops an iterative reasoning framework that integrates an LLM, a multifunctional toolbox, a KG-based executor, and a knowledge memory module, enabling smaller LLMs to make decisions autonomously. 
SymAgent~\cite{liu2025symagent} conceptualizes the KG as a dynamic environment and is designed to autonomously and effectively integrate the capabilities of both the LLM and the KG.
Graph-R1~\cite{luo2025graph} proposes lightweight knowledge hypergraph construction and retrieval framed as a multi-turn agent–environment interaction, optimizing the reasoning process via an end-to-end reward mechanism. DynaSearcher~\cite{hao2025dynasearcher} utilizes knowledge graphs as external structured knowledge to guide the search process by explicitly modeling entity relationships, thereby ensuring factual consistency in intermediate queries.}

\section{Methodology}

In this section, we introduce Graph-RFT, with autonomous planning and adaptive retrieval capabilities for reasoning over incomplete KGs.
The framework is structured around two complementary core stages: (1) CoT Fine-Tuning for Reasoning Activation: 
we propose a supervised fine-tuning (SFT) method using high-quality CoT plan–retrieval trajectories to activate the model’s structured reasoning and planning abilities while resolving the GRPO cold-start problem. 
(2) RL-based Reasoning Enhancement: we propose a plan–retrieval guided RL method with a multi-reward design to optimize reasoning and retrieval coordination, enabling the model to perform coherent multi-step planning and adaptively invoke KG and external tools under incomplete knowledge conditions.
We start with the definition of the KGQA task.

\subsection{Task Formulation}
\begin{sloppypar}
\vic{Let  $D$  be a dataset containing question–answer pairs $(q,a)$, $\mathcal G$ a KG  containing a set of triples
$ \{ (e_i, r, e_j) \ | e_i,e_j \in \mathcal{E}; r \in \mathcal{R} \}$, where $\mathcal{E}$ is a set of entities and $\mathcal{R}$ is the set of relations names,  and $E$ an external search engine. The KGQA task requires the LLM to generate reasoning trajectories or  to iteratively interact with $\mathcal G$ and  $E$. Formally, for each question $q$, we generate a reasoning trajectory: $o = (\tau_1, \tau_2, . . . , \tau_T )$, where the $t$-th intermediate reasoning step $\tau_t = (s_t, c_t)$ consists of an action $s_t \in \{\langle think\rangle, \langle plan\rangle,$ $ \langle relation\_search\rangle, \langle neighbor\_search\rangle, \langle web\_search\rangle, \langle answer\rangle\}$ (cf. Figure~\ref{fig:prompt}) and its associated content $c_t$. The model is expected to repeatedly retrieve knowledge from $\mathcal G$ and $E$ until reaching the final answer $a$  for the question $q$.}

\end{sloppypar}


\subsection{CoT Fine-Tuning for Reasoning Activation}
\label{data_collection}
\begin{figure*}[t!]
    \centering
    \includegraphics[width=.9\textwidth]{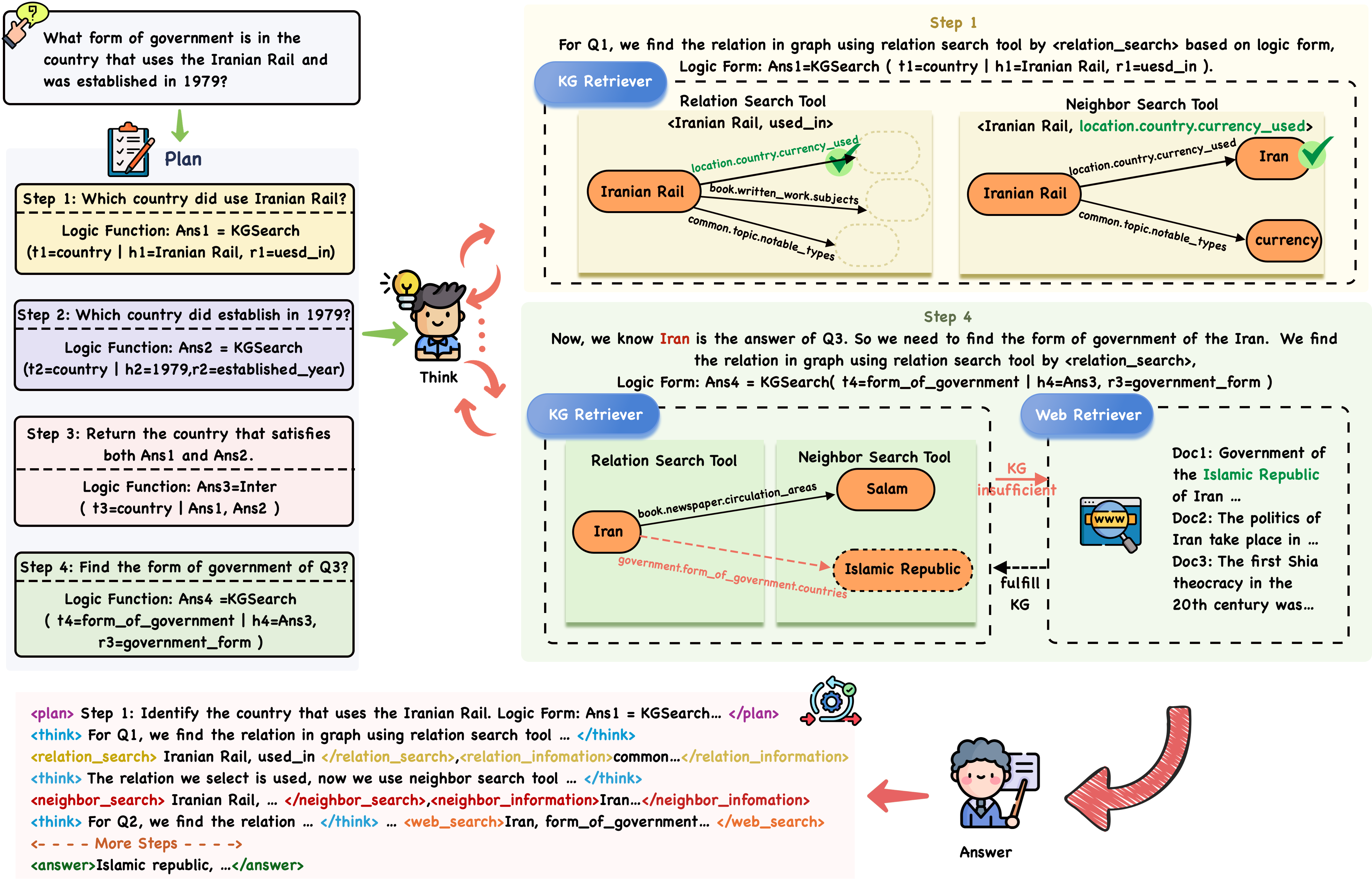}
    \caption{Graph-RFT Reasoning Trajectory for SFT dataset construction and rollout process for RL.}
    \label{fig:dataset}
    \vspace{-2ex}
\end{figure*}

\vic{In the initial phase, we propose the CoT finetuning method using a structured reasoning dataset that contains step-by-step planning, reasoning, and retrieval processes. This phase is critical for activating the model's planning-reasoning capabilities and establishing a robust cold start for subsequent RL.
To construct our SFT dataset, as illustrated in Figure \ref{fig:dataset}, we monitor and record the agent's output throughout a multi-step interactive process with the KG and the external search engine. This approach transforms the complex, multi-step collaboration procedure into a  CoT-style trajectory suitable for SFT. In addition, we use a two-stage process to filter the trajectory to get the high quality dataset. The Details regarding the data's annotation and filtering processes are provided in \textcolor{blue}{Appendix \ref{app:data_construct}}. The training objective minimizes:}
\begin{equation}
    \mathcal{L}_{\text{SFT}} = - \sum_{t=1}^{T} \log \pi_{\theta}(y_t \mid q, y_{<t})
\end{equation}
\vic{where $q$ is the question, $y$ is the concatenated sequence of reasoning steps and the answer, $\pi_{\theta}$ is the model's token distribution.
The output model $\pi_{CoT}$ serves as the initialization for the next stage \cite{tan2025reason}, ensuring a robust foundation for RL.}

\subsection{RL-based Reasoning Enhancement}
\label{second_stage}
\subsubsection{Rollout with Search Tools}
\label{rollout}

\vic{Our approach follows an iterative framework where the LLM alternates between text generation and external search engine queries. Specifically, our method consists of three steps: 
i) Plan the sub-problem decomposition and tool invocation for complex reasoning in a specific order; 
ii) Retrieve corresponding answers from the knowledge graph based on the priority and logical functions of these subproblems;
iii) Obtain the correct answers by searching external sources for information not available in the knowledge graph.}

\noindent \textbf{Plan the Steps.} 
\vic{We adhere to the Cartesian principle of breaking down complex problems into distinct steps, while designing logical functions to assist the model in understanding the interrelationships between sub-problems and determining the appropriate timing for tool utilization.
In Figure \ref{fig:dataset}, the first two steps correspond to subproblems of higher priority, as they do not rely on answers from other subproblems. Additionally, we use logical functions to represent the parameters for invoking KG retrieval tool, which helps the model gain a clear understanding of the invocation process. Step 3 expresses the logical relationship between the two subproblems through the $\textit{inter}(Ans_1, Ans_2, .., Ans_n)$  function (the intersection of the answers to subquestions), passes the returned results to Step 4, and obtains the final answer by invoking the KG retrieval tool.
The planning steps are encapsulated between <plan> and </plan>.} 

\noindent \textbf{Retrieve Facts from the KG.} \vic{For each single steps obtained through planning, we need to sequentially select relevant triples from the KG. To address this, we design two interactive toolkits.}

\begin{itemize}[itemsep=0.5ex, leftmargin=3mm]
\item \textbf{Relation Search Tool.} \vic{This tool retrieves a set of candidate relations for an entity in a given single-hop problem. The LLM encapsulates the entity and the hypothetical relation from the logical function between the relation search tokens <relation\_search> and </relation\_search>. By invoking the relation acquisition toolkit, the set of top 15 relations that are similar to the hypothetical relation is encapsulated between the tokens <relation\_information> and </relation\_information>.}
\item \textbf{Neighbor Search Tool.} \vic{This tool retrieves the tail entity corresponding to a given head entity-relation  pair selected by LLMs. The LLM encapsulates the specified head entity and relation between the neighbor search tokens <neighbor\_search> and </neighbor\_search>. By invoking the neighbor acquisition toolkit, the answer is then encapsulated between the tokens <neighbor\_information> and </neighbor\_information>.}

\end{itemize}
\vic{Given the obtained single-hop questions (e.g., ``Identify the country that uses the Iranian Rail''), we first invoke the Relation Search Tool to retrieve a candidate set of relations associated with the initial (topic) entity. The LLM then reasons over this set and selects the most appropriate relation, for example, choosing ``location.country.currency\_used'' to replace ``used\_in'' in the retrieval function.
This greedy strategy simplifies the reasoning process by selecting a locally optimal relation at each step, thereby reducing reliance on global contextual information. Next, the Neighbor Search Tool is called to obtain specific triples and their corresponding answers (e.g., ``Iran'').
If the LLM determines that the retrieved triples are insufficient to answer the question, or if no answer is found in the knowledge graph (i.e., when <neighbor\_information> and </neighbor\_information> return “No information in the KG, please use web tool”), the system directly invokes the search engine.
The answers to higher-priority (according to the order produced by the subquestion plan)
single-hop questions can then be combined with those of lower-priority ones to generate new single-hop questions. By iteratively repeating this process, the system is able to derive answers to complex multi-hop questions. Notably, if the maximum iteration limit is reached without successfully generating an answer, the parametric knowledge of the LLM is used to provide a final response.}

\vic{When LLMs determine that the knowledge retrieved through KG search tools is insufficient or implausible (such as when the KG lacks relevant information, i.e., the tags <neighbor\_information> and </neighbor\_information> return ``No information in KG, please use web tool.''), the model encapsulates the corresponding head entity and relation, for which no information is available, within the special search tokens <web\_search> and </web\_search>. Subsequently, a search engine is invoked, and the documents retrieved from the Bing website are enclosed within <web\_information> and </web\_information>. This mechanism not only enables the model to acquire the supplementary knowledge required to answer the query but also facilitates the identification of missing  triples, which can later be incorporated into the KG to mitigate its incompleteness. In our experiments, we fixed the top-$k$ value (e.g., 3) to control the number of retrieved documents.
}
\subsubsection{Reward Design}
\label{rewards}
\begin{sloppypar}
    To further optimize the above retrieval-planning process, we propose the GRPO \cite{shao2024deepseekmath} based RL method with the novel multi-reward design. Prior studies have predominantly employed outcome-based rewards to guide LLMs in developing reasoning abilities and leveraging search engines throughout the reinforcement learning process. However, such supervision alone does not sufficiently capture diverse retrieval-oriented reward mechanisms and is inadequate for effectively guiding retrieval in the presence of incomplete KGs. To address this limitation, we incorporate two key components into our reward design: (i) an outcome-based reward, which directly evaluates the quality of the model’s generated answers; and (ii) a retrieval-specific reward, which encourages the model to discern when and how to extract relevant information from knowledge graphs or web documents.
\end{sloppypar}

\begin{table*}[t]
\caption{Results of Graph-RFT across various datasets (represent the maximum hop), compared with the state-of-the-art (SOTA) in Supervised Learning (SL) and In-Context Learning (ICL) methods. The highest scores for ICL methods are highlighted in bold, while the second-best results are underlined. The Prior FT (Fine-tuned) SOTA includes the best-known results achieved through supervised learning. CKG denotes using complete knowledge graph and IKG denotes using incomplete KG (IKG-40\%).}
\label{fig:mainresult}
\small
\begin{tabular}{@{}ccccccccccc@{}}
\toprule
Method        & Class & LLM           & \multicolumn{6}{c}{Multi-Hop KGQA}   & \multicolumn{2}{c}{Single-Hop KGQA} \\ \cmidrule(lr){4-9}
              &       &               & \multicolumn{2}{c}{CWQ (4-hop)}  & \multicolumn{2}{c}{WebQSP (2-hop)}        & \multicolumn{2}{c}{GrailQA (4-hop)}       & \multicolumn{2}{c}{Simple Questions (1-hop)}    \\ \midrule
\multicolumn{11}{c}{\textit{Without external knowledge}}                                                                   \\\midrule
IO prompt\cite{sun2023think}     & -     & GPT-3.5-Turbo & \multicolumn{2}{c}{37.6} & \multicolumn{2}{c}{63.3}          & \multicolumn{2}{c}{29.4}          & \multicolumn{2}{c}{20.0}                      \\
CoT\cite{sun2023think}            & -     & GPT-3.5-Turbo & \multicolumn{2}{c}{38.8} & \multicolumn{2}{c}{62.2}          & \multicolumn{2}{c}{28.1}          & \multicolumn{2}{c}{20.3}                     \\
SC\cite{sun2023think}             & -     & GPT-3.5-Turbo & \multicolumn{2}{c}{45.4} & \multicolumn{2}{c}{61.1}          & \multicolumn{2}{c}{29.6}          & \multicolumn{2}{c}{18.9}                     \\ \midrule
 &       &               & CKG  & IKG  &  CKG & IKG       & CKG   & IKG       & CKG  & IKG   \\ \midrule
\multicolumn{11}{c}{\textit{With external knowledge}}                                                                      \\ \midrule
RoG  \cite{luo2023reasoning}  & SL    & Qwen2.5-7B-base        & 63.7 & 53.6 & 84.2 & 75.3        & -   & -     & -    & -           \\
ChatKBQA \cite{luo2023chatkbqa}  & SL     & Qwen2.5-7B-base      & 72.4  & 37.8 & 75.5      & 47.1    & -   & -     & -   & - \\ 
\midrule
KB-BINDER\cite{li2023few}     & ICL   & Codex         & -  & -  & 50.7 & 38.4          & -      & -    & -  & -                     \\
StructGPT\cite{jiang2023structgpt}     & ICL   & GPT-3.5-Turbo & - & - & 76.4    & 60.1      & -     & -     & -    & -  
\\
ToG/ToG-R\cite{sun2023think}     & ICL   & GPT-3.5-Turbo         & 47.2 & 37.9 & 76.9 & 63.4          & -    & -      & -    & -                  \\ 
ToG/ToG-R\cite{sun2023think}     & ICL   & GPT-4         & 71.0 & 56.1 & 80.3 & 71.8    & -  & -        & -    & -              \\ 
PoG\cite{tan2025paths}     & ICL   & GPT-4        & 76.8 & 62.6 & \textbf{96.7} & 76.4   & 79.2  & 64.1       & \textbf{80.2}   & \underline{59.5}            \\ 
GoG\cite{xu2024generate}   & ICL   & GPT-3.5-Turbo         & 55.7 & 44.3 & 78.7  & 66.6         & 71.7  & 62.4        & 54.3   &  43.9                 \\ 
GoG\cite{xu2024generate}  & ICL   & GPT-4         & 75.2 & 60.4 & 84.4     & 80.3     & 76.8   & 69.4       & 68.9   & 56.4                \\ 
\midrule

Graph-RFT-instruct         &  RL  & Qwen2.5-7B-instruct & \underline{78.4}   & \underline{64.8}          &  87.3         & \underline{82.7} &\underline{82.1} &\underline{71.8} &73.8 & 58.7                \\
Graph-RFT-base         & RL   & Qwen2.5-7B-base &  \textbf{80.7}  &\textbf{67.2}           &   \underline{90.6}       &  \textbf{86.3} &\textbf{84.6} &\textbf{73.3} &\underline{76.9} &\textbf{62.4}                   \\ \bottomrule
\end{tabular}
\end{table*}

\noindent \textbf{Outcome-based Reward.}
\vic{Result-based reward are divided into format reward and answer reward. For format reward, we first define the correct format to ensure that the LLM adopts the predefined iterative workflow of "Think-Decompose-Retrieval-Search". The definitions of the model's thinking process, tool invocation, and final answer output format are shown in Figure \ref{fig:prompt} in \textcolor{blue}{Appendix \ref{app:prompt}}.
For answer rewards $r_{ans} \in [0, 1]$, we compare the predicted answer in  <answer></answer> with the ground-truth answer, and use the F1-score between the two sets as the reward to measure its accuracy. The complete accuracy reward $R_{acc}$ is defined as:}
\begin{equation}
    R_{\text{acc}} = 
\begin{cases} 
\max(0.1, r_{\text{ans}}), & \text{if format is correct}, \\
0, & \text{if format is incorrect}.
\end{cases} 
\end{equation}

\begin{equation}
r_{\text{ans}} = \text{F1}(p_{\text{ans}}, a) = \frac{2|p_{\text{ans}} \cap \{a\}|}{|p_{\text{ans}}| + |a|}
\end{equation}
\vic{where \( p_{\text{ans}} \) is the predicted answer, and \( a \) is the ground truth answer from the \((q, a)\) pair.}

\noindent \textbf{Retrieval-specific Reward.}
\vic{Inspired by AutoRefine~\cite{shi2025search}, building on the result-based reward described above, we further introduce two retrieval rewards to encourage LLMs to learn how to extract relevant information from both the KG and web documents. Specifically, we introduce a  graph retrieval reward $R_{graph}$ and a document search reward $R_{web}$. }

\begin{sloppypar}
\vic{$R_{graph}$ is measured based on the retrieval results contained in the <neighbor\_information></neighbor\_information> block. Specifically, we collect all graph retrieval results from the entire trajectory and concatenate them into a single text sequence:}
\begin{equation}
    \mathcal{R}_{\text{graph}} = \mathbb{I}\bigl(\{a\} \cap o_{\text{graph}} = a\bigr)
\end{equation}
\begin{equation}
    o_{\text{graph}} = \bigcup \left\{ c_t \mid (s_t, c_t) \in o \land s_t = \text{<neighbor\_information>} \right\}
\end{equation}

\vic{where $I(\cdot)$ is the indicator function, $o_{graph}$ is the concatenation of all the KG retrieval information steps.}

\vic{$R_{web}$ is defined in a similar way to $R_{graph}$. We collect all document retrieval results encapsulated in the <web\_information></web\_information> blocks throughout the entire trajectory and concatenate them into a single text sequence.}
\begin{equation}
    \mathcal{R}_{\text{web}} = \mathbb{I}\bigl(\{a\} \cap o_{\text{web}} = a\bigr)
\end{equation}
\begin{equation}
    o_{\text{web}} = \bigcup \left\{ c_t \mid (s_t, c_t) \in o \land s_t = \text{<web\_information>} \right\}
\end{equation}

\vic{where $I(\cdot)$ is the indicator function, $o_{graph}$ is the concatenation of all the web search information steps.} 
    
\vic{However, such multi-turn reinforcement learning is unstable. To prevent the model from bypassing graph retrieval and directly turning to document search in cases where the correct result could have been obtained through graph retrieval, we introduce a corresponding penalty for this behavior to help LLMs know when to use different retrieval tools. The overall reward $\mathcal R_{\text{over}}$ can be formally written as:}
\begin{small}
\begin{equation}
    \mathcal{R}_{\text{over}} = 
\begin{cases} 
\mathcal{R}_{\text{acc}}, & \text{if } \mathcal{R}_{\text{acc}} > 0, \\
0.1, & \text{if } \mathcal{R}_{\text{acc}} = 0 \text{ \& } \mathcal{R}_{\text{graph}} > 0   \text{ OR }     \mathcal{R}_{\text{acc}} = 0 \text{ \& }  \mathcal{R}_{\text{web}} > 0,\\
-0.1, & \text{if } \mathcal{R}_{\text{web}} > 0 \text{ when CKG} \text{ OR } \mathcal{R}_{\text{web}} = 0 \text{ when IKG}\\
0, & \text{if } \mathcal{R}_{\text{acc}} = \mathcal{R}_{\text{web}} = \mathcal{R}_{\text{graph}} = 0.
\end{cases}
\end{equation}
\end{small}
\end{sloppypar}

\noindent \textbf{Training Objective.}
\vic{By integrating the overall reward with the GRPO training objective~\cite{shao2024deepseekmath}, our proposed learning objective enables dynamic interaction with both the knowledge graph and the search engine during optimization~\cite{jin2025search}, thereby facilitating effective LLM search training.
The objective is formalized as:}
\begin{equation}
\begin{split}
\max_{\pi_\theta} \mathbb{E}_{x \sim \mathcal{D}, \, y \sim \pi_\theta(\cdot | x; \mathcal{G},\mathcal{R})} \left[ r_\phi(x, y) \right] - \\ \beta \mathbb{D}_\text{KL} \left[ \pi_\theta(y | x; \mathcal{G},\mathcal{R}) \big\| \pi_\text{ref}(y | x; \mathcal{G},\mathcal{R}) \right]
\end{split}
\end{equation}
\vic{where $\pi_\theta$ denotes the trainable policy model, $\pi_\text{ref}$ is a fixed reference model, $r_\phi$ represents the overall reward function, and $\mathbb{D}_\text{KL}$ denotes the $KL$ divergence. Here, $x$ is sampled from the dataset $D$, and $y$ denotes the output sequence interleaving reasoning steps with KG and search engine retrievals. Since the retrieved triples and documents are not generated by the policy model, we mask the retrieval results during loss calculation to prevent the training policy from being biased \cite{hao2025dynasearcher}.}

\section{Experiments}
To evaluate the effectiveness of Graph-RFT, we aim to explore the following four research questions (RQs):
(1) \textbf{RQ1}: How does Graph-RFT perform against SOTA baselines on complex reasoning datasets?
(2) \textbf{RQ2}: What are the contributions of each  Graph-RFT module to overall performance?
(3) \textbf{RQ3}: Can Graph-RFT effectively bridge information gaps via retrieval under varying KG incompleteness?
(4) \textbf{RQ4}: What are the error types of Graph-RFT?

\begin{sloppypar}
\vic{\noindent\textbf{$\blacktriangleright$ Datasets.} We evaluate Graph-RFT on four KGQA datasets, including three multi-hop datasets: CWQ~\citep{talmor2018web}, WebQSP~\citep{yih2016value}, GrailQA~\citep{gu2021beyond} and one single-hop dataset: SimpleQuestions~\citep{petrochuk1804simplequestions}. We consider two scenarios: Complete Knowledge Graph (CKG) and Incomplete Knowledge Graph (IKG). For CKG, we directly use the datasets provided by GoG~\citep{xu2024generate}. For IKG, since GoG only provides CWQ and WebQSP, we construct IKG versions of GrailQA and SimpleQuestions by sampling 3,000 multi-hop questions from GrailQA and 3,000 questions from SimpleQuestions. We further generate four IKGs with varying completeness levels IKG-20\%, IKG-40\%, IKG-60\%for example, IKG-40\% removes 40\% of the critical triples for each question and all relations between the corresponding entity pairs. Freebase~\citep{bollacker2008freebase} serves as the background KG for all datasets. 
The impact of using alternative knowledge graphs is discussed in \textcolor{blue}{Appendix~\ref{knowledege_bases}}.
}\\
\vic{\noindent\textbf{$\blacktriangleright$ Evaluation Metrics.} Following prior work~\cite{li2023chain, baek2023knowledge, jiang2023structgpt, sun2023think}, we adopt exact match accuracy (Hits@1) as the evaluation metric for all datasets.
Detailed descriptions of the baselines and implementation settings are provided in \textcolor{blue}{Appendix~\ref{app:baseline}} and \textcolor{blue}{Appendix~\ref{app:imp}}.
}
\end{sloppypar}

\subsection{\zhiwei{Main Results (RQ1)}}

\vic{Table \ref{fig:mainresult} presents our main results on four KGQA datasets. We can observe that Graph-RFT consistently achieves SOTA performance across all settings in IKG, even when built upon smaller backbones such as Qwen2.5-7B-base, outperforming GPT-4-based methods.}

\noindent  $\blacktriangleright$ \textbf{CKG Setting.} Under the CKG setting, where all relevant triples are available, Graph-RFT achieves the strongest overall performance across both multi-hop and single-hop benchmarks. The results reveal that our two-stage RFT which combines CoT-based reasoning activation and plan–retrieval RL—provides consistent advantages over both prompt-only ICL methods and supervised learning baselines.
ICL approaches such as PoG and GoG rely purely on prompt engineering and lack any fine-tuning or reinforcement optimization. Although they perform competitively on simpler datasets such as WebQSP and SimpleQuestions, they struggle to generalize across long reasoning chains. In contrast, Graph-RFT-base (Qwen-2.5-7B) achieves substantial gains on complex benchmarks like CWQ (80.7 vs 76.8/75.2) and GrailQA (84.6 vs 79.2/76.8), where multi-step logical dependencies are required. These improvements highlight how explicit CoT supervision enables structured planning, and reinforcement optimization refines the timing and sequencing of retrieval actions.
Relative to RoG and ChatKBQA, which are purely supervised fine-tuned models, Graph-RFT demonstrates marked improvements (e.g., CWQ 80.7 vs 72.4/63.7, WebQSP 90.6 vs 84.2/75.5). This suggests that reinforcement learning contributes beyond imitation: by rewarding accurate retrieval scheduling and penalizing inefficient queries, the model learns to maintain global reasoning consistency and to balance exploration between relation-search and neighbor-search tools.

\begin{table}[htbp]
\centering
\caption{Ablation study for the Graph-RFT-base in rollout framework.}
\resizebox{\linewidth}{!}{
\label{tab:ablation}
\begin{tabular}{l|ccc|cccc}
\toprule
&\textbf{PS} & \textbf{KR} & \textbf{WR}  & \multicolumn{3}{c}{\textbf{Multi-Hop KGQA}}   & \multicolumn{1}{c}{\textbf{Single-Hop KGQA}} \\ \cmidrule(lr){5-7}
 & &  &  & CWQ & WebQSP & GrailQA & Simple Questions \\
\midrule
 \textbf{Variants}& -& \checkmark& -  & 49.4 & 69.7 & 58.9 & 53.8 \\
  & - & \checkmark  &\checkmark  & 55.3 & 74.2 & 63.5 & \underline{61.7} \\
 & \checkmark &\checkmark  & - & \underline{62.6} & \underline{83.4} & \underline{69.1} & 57.3 \\ \midrule
\textbf{Graph-RFT-base} & \checkmark & \checkmark  & \checkmark & \textbf{67.2} & \textbf{86.3} & \textbf{73.3} & \textbf{62.4} \\
\bottomrule
\end{tabular}
}
\end{table}

\begin{table}[htbp]
\centering
\caption{Ablation analysis on Graph-RFT-base SFT and multi-reward design.}
\resizebox{\linewidth}{!}{
\label{tab:reward}
\begin{tabular}{l|cccc}
\toprule
Method & \multicolumn{3}{c}{\textbf{Multi-Hop KGQA}}   & \multicolumn{1}{c}{\textbf{Single-Hop KGQA}} \\ \cmidrule(lr){2-4}
   & CWQ   & WebQSP   & GrailQA & Simple Questions \\
\midrule
Graph-RFT-base w/o SFT  &46.4&74.2& 56.4 &54.7 \\
Graph-RFT-base w/o $R_{web}$ & 66.3 & 83.5  &72.1 & 60.8 \\
Graph-RFT-base w/o $R_{graph}$ & 65.8& 82.7 & 71.6 & 61.6\\ \midrule
\textbf{Graph-RFT-base} & \textbf{67.2} & \textbf{86.3}  & \textbf{73.3} & \textbf{62.4}\\
\bottomrule
\end{tabular}
}
\end{table}

 \noindent $\blacktriangleright$   \textbf{IKG Setting.} 
When the knowledge graph is incomplete, the benefits of Graph-RFT become even more pronounced. 
Prompt-based ICL approaches such as PoG and GoG (GPT-4) achieve competitive accuracy under complete KGs but degrade significantly once KG coverage is reduced. For example, GoG’s performance drops 7-15 points across CWQ (75.2→60.4) and GrailQA (76.8→69.4), while PoG falls from 76.8→62.6 on CWQ. These methods rely on static prompt patterns and local reasoning, lacking mechanisms to assess KG sufficiency or to seek supplementary evidence.
In contrast, Graph-RFT-base (Qwen2.5-7B) remains stable across both settings, outperforms among ICL methods. This resilience arises from our two-stage RFT: CoT fine-tuning teaches explicit question decomposition, and reinforcement optimization learns when and how to invoke external retrievals.
SL methods such as RoG and ChatKBQA also decline sharply under IKG (e.g., CWQ: 63.7→53.6; ChatKBQA: 72.4→37.8), as their fixed fine-tuning lacks interactivity with retrieval tools. In contrast, Graph-RFT maintains high accuracy (e.g., CWQ 67.2, WebQSP 86.3), confirming that reinforcement-based retrieval scheduling effectively compensates for missing triples through controlled web augmentation.
Moreover, Graph-RFT with Qwen2.5-7b-base outperforms methods without external knowledge (e.g., IO, CoT, SC prompting), demonstrating that KGs play a crucial role in reasoning tasks.
\subsection{\zhiwei{Ablation Studies (RQ2)}}


\zhiwei{We perform ablation studies to analyze the contributions of Graph-RFT’s key components. Specifically, we examine the effects of the rollout framework, SFT and the multi-reward mechanism in the RL stage.}\\
\zhiwei{\noindent\textbf{$\blacktriangleright$ Exploration on Rollout Framework.} We analyze the contributions of the Planning Steps (PS), KG Retrieval (KR), and Web Retrieval (WR) modules within the rollout framework to Graph-RFT’s performance under the IKG-40\% setting. Figure~\ref{tab:ablation} presents model variants obtained by ablating these components, revealing that removing any single module substantially reduces performance. Comparing PS+KR, KR+WR, and KR alone, we observe that PS+KR achieves superior performance on multi-hop datasets, indicating that global planning—including question decomposition with logical operators and tool invocation—provides greater benefit than merely supplementing missing knowledge, thereby validating the effectiveness of the PS module. In contrast, on the single-hop dataset, KR+WR outperforms PS+KR, suggesting that leveraging external unstructured knowledge becomes more critical than global planning for simpler queries, which demonstrates Graph-RFT’s adaptability across tasks of varying complexity. Furthermore, the comparison between KR+WR and KR alone confirms that integrating structured knowledge from the KG with unstructured knowledge from web documents produces substantial improvements. Collectively, these results indicate that each module contributes uniquely, and their synergistic combination significantly enhances the model’s capability for complex reasoning tasks.
}
\begin{figure}[t]
  \centering
  \begin{tabular}{@{}cc@{}}
    \includegraphics[width=.5\linewidth]{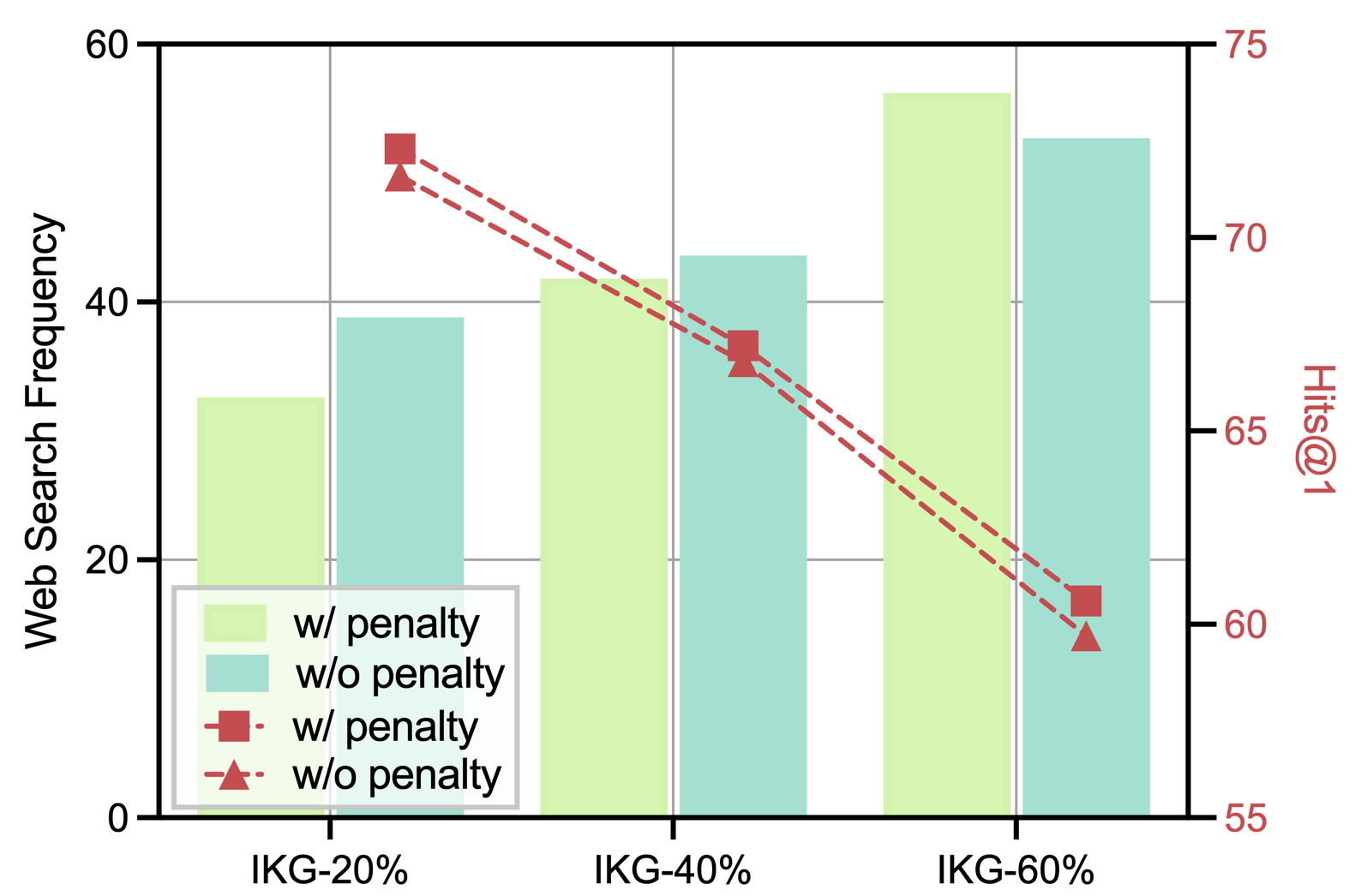} &
    \includegraphics[width=.5\linewidth]{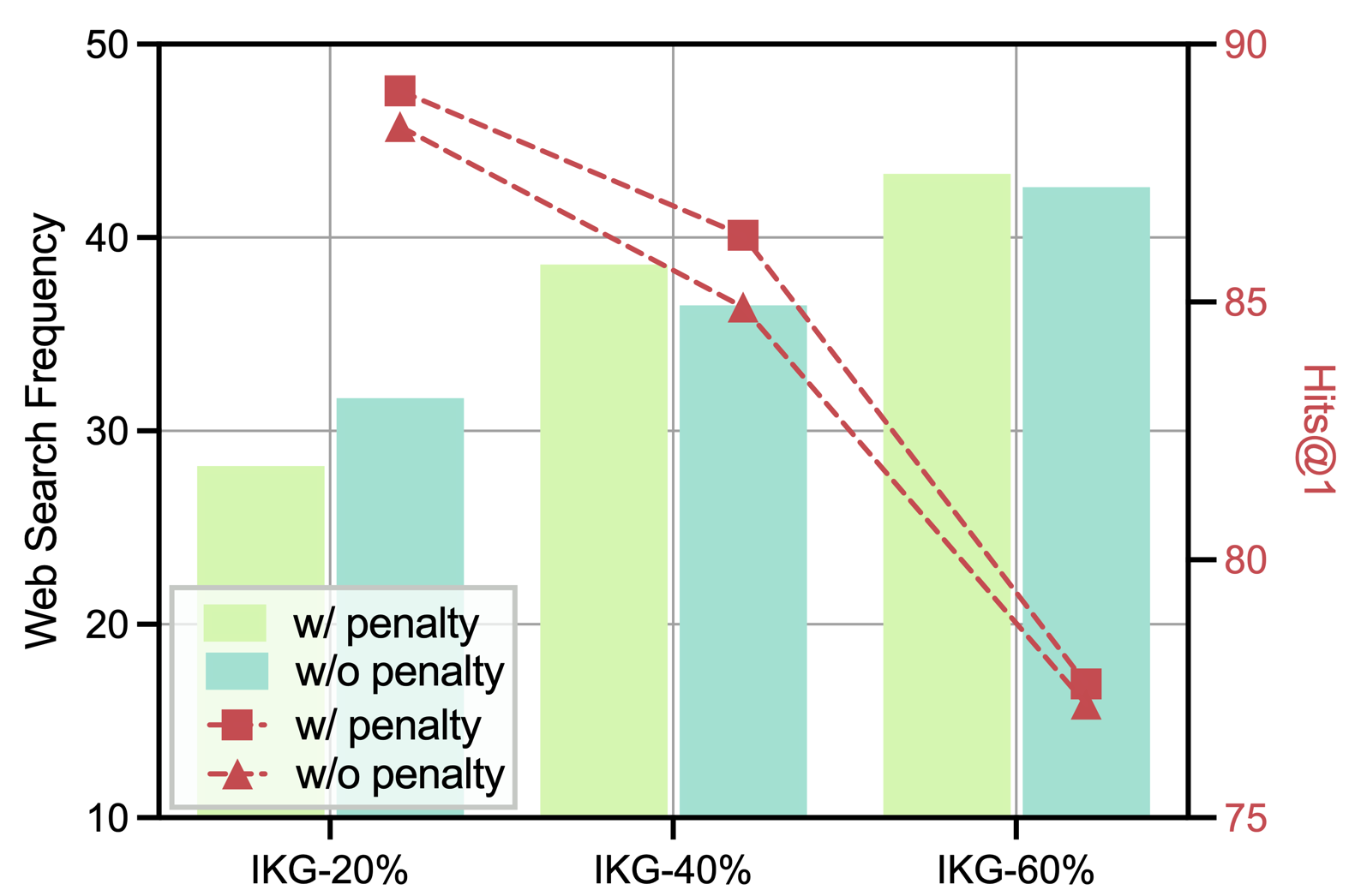}
    \\
    \textbf{(a) CWQ.} &
    \textbf{(b) WebQSP.}
  \end{tabular}
  \caption{Ratio of web search operation in different KG settings under rewards with penalty or not.}
  \label{fig:exp2}
\end{figure}
\begin{figure}[t]
  \centering
  \begin{tabular}{@{}cc@{}}
    \includegraphics[width=.5\linewidth]{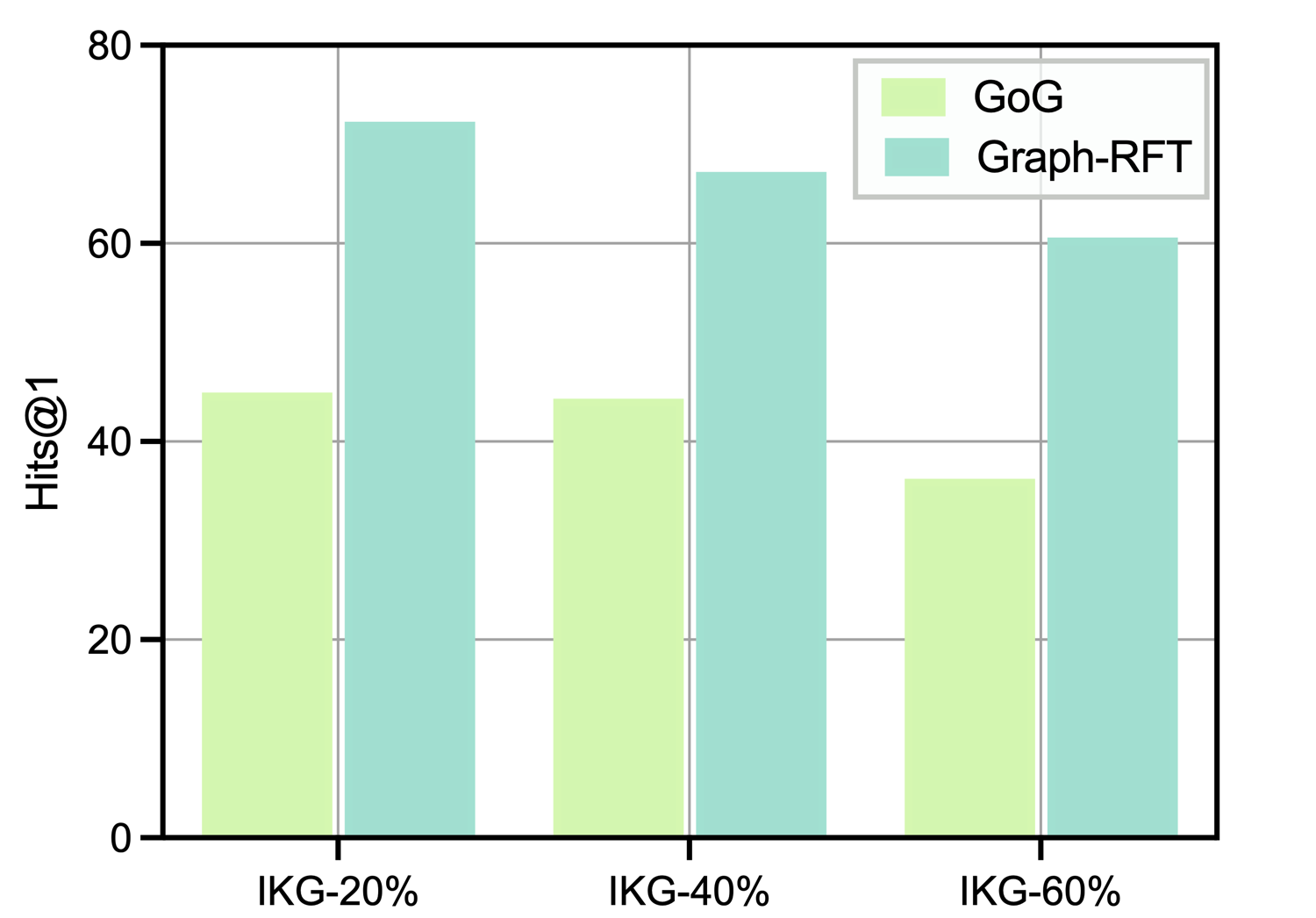} &
    \includegraphics[width=.5\linewidth]{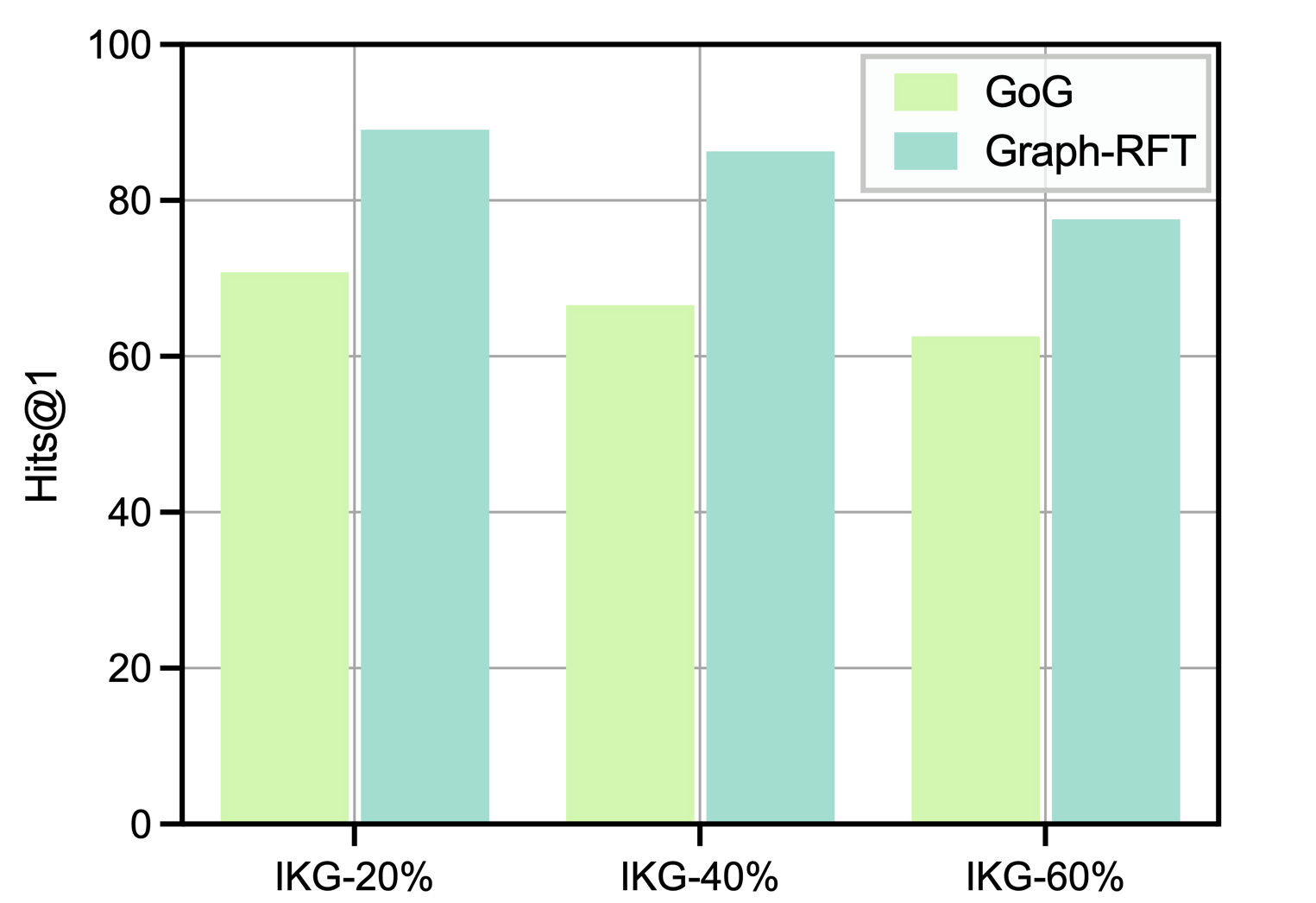}
    \\
    \textbf{(a) CWQ.} &
    \textbf{(b) WebQSP.}
  \end{tabular}
  \caption{Performance of GoG and our model under different KG settings.}
  \label{fig:compare}
\end{figure}
\\\zhiwei{\noindent\textbf{$\blacktriangleright$ Exploration on SFT and Multi-reward Design.} We further examine the effect of the SFT stage and the multi-reward mechanism on Graph-RFT’s performance under the IKG-40\% setting. Comparing Graph-RFT-base w/o SFT with Graph-RFT-base shows a marked performance drop across all datasets, underscoring the crucial role of SFT in strengthening model capability. Moreover, removing either the web retrieval reward or the graph retrieval reward consistently degrades performance, demonstrating that both rewards effectively guide the LLM to retrieve more semantically relevant triples and external documents. As shown in Figure~\ref{fig:exp2}, when the model’s web search frequency approaches the optimal level (e.g., 20\% in the IKG-20\% setting), performance on CWQ and WebQSP improves. This indicates that Graph-RFT can adaptively regulate web search frequency based on the sufficiency of KG information, reducing the risk of noisy or irrelevant content and thereby enhancing overall effectiveness.
}

\subsection{\zhiwei{Different KG Incompleteness (RQ3)}}
\label{exp2}

\zhiwei{The key capabilities of our model lies in performing global planning on complex problems to leverage the knowledge within the KG and identifying knowledge gaps beyond the KG through web search. We examine the performance of Graph-RFT under varying degrees of KG incompleteness and record its external search frequency, as shown in Figure~\ref{fig:exp2}. To evaluate the robustness of different methods under KG incompleteness, we compare GoG and Graph-RFT across KGs with different completeness levels, as illustrated in Figure~\ref{fig:compare}.
}\\ 
\zhiwei{\noindent\textbf{$\blacktriangleright$ Impact of Rewards with or without Penalty.} From Figure~\ref{fig:exp2}, Graph-RFT increases the frequency of external searches as KG completeness decreases. This behavior reflects the model’s ability to proactively compensate for missing knowledge required for reasoning through web retrieval. Moreover, the model’s search strategy varies notably with question complexity: compared to WebQSP (maximum 2-hop), Graph-RFT conducts significantly more external searches when tackling the more complex multi-hop questions in CWQ (maximum 4-hop). This demonstrates that the model can dynamically adjust its web search frequency according to question complexity, effectively aligning knowledge acquisition with the reasoning demands of the task. }\\
\zhiwei{\noindent\textbf{$\blacktriangleright$ Impact of Different KG Incompleteness.} From Figure~\ref{fig:compare}, we observe that Graph-RFT outperforms GoG across all degrees of incompleteness. This improvement arises because GoG is primarily confined to local decision-making, lacking the global planning needed for effective question understanding and tool invocation in complex problem decomposition. In contrast, when KG triples are missing, GoG relies solely on the LLM’s internal knowledge to generate answers, whereas our model retrieves relevant external documents through web search, thereby supplementing incomplete KG information and improving answer accuracy. Notably, the improvement on CWQ is much more pronounced than on WebQSP, validating Graph-RFT’s strong planning ability and logical decomposition mechanism for tackling complex reasoning tasks.}

\subsection{\zhiwei{Error Type Analysis (RQ4)}}
\zhiwei{To analyze Graph-RFT’s limitations, we randomly selected 50 failure cases from CWQ and WebQSP under both CKG and IKG settings. These cases are categorized into five types: (1) invalid actions, (2) relation selection errors, (3) neighbor selection errors, (4) reasoning errors, and (5) decomposition errors, with their distribution shown in Figure~\ref{fig:error}. We can obtain the following findings: (\textit{i}): reasoning errors—where the LLM follows a correct reasoning process but produces incorrect answers—represent the largest proportion in both datasets and are more prevalent in IKG than in CKG, likely due to answer aliasing and discrepancies between retrieved document answers and reference answers. (\textit{ii}): neighbor selection errors manifest in two scenarios: on the one hand, when relation selection is correct, neighbor errors are relatively rare in CKG, indicating that the model primarily struggles with relation selection; on the other hand, when the correct relation is not identified, as in IKG, the model often fails to determine whether entities are sufficient for problem-solving, resulting in incorrect neighbor selection. (\textit{iii}): decomposition errors, arising from flawed planning, occur more frequently on complex problems such as CWQ, whereas invalid action errors (e.g., invoking undefined tools) are effectively managed by the model. The case study is in \textcolor{blue}{Appendix \ref{case_study}}}.

\section{Conclusion}
We introduced Graph-RFT, a two-stage reinforcement fine-tuning framework for complex reasoning over incomplete knowledge graphs. Graph-RFT unifies structured planning and adaptive retrieval within a single learning paradigm, enabling LLMs to reason coherently and retrieve information dynamically across KGs and web sources.
In the first stage, a CoT fine-tuning process with a customized plan–retrieval dataset activates the model’s planning and reasoning capabilities while mitigating the GRPO cold-start problem. In the second stage, a plan–retrieval guided reinforcement learning procedure with a multi-reward design optimizes coverage-aware retrieval scheduling and multi-step logical consistency.
Extensive experiments across multiple KGQA benchmarks demonstrate that Graph-RFT consistently outperforms strong baselines, achieving superior accuracy even with smaller LLM backbones.
In the future, we will explore enhancing relation
filtering performance from KGs and the capability of decomposing problems when dealing with complex reasoning tasks.

\begin{acks}
We thank all anonymous reviewers and area chairs for their comments. This work is supported by the National Natural Science Foundation of China (U23A20316), the CCF-Tencent Rhino-Bird Open Research Fund (CCF-Tencent RAGR20250115), and the computing power resources from Wuhan Dongxihu District Intelligent Computing Center.
\end{acks}

\bibliographystyle{ACM-Reference-Format}
\balance
\bibliography{sample-base}

@String{Computing = "Computing" }

@String{Academic = "Academic Press" }

@String{Springer = "Springer-Verlag" }

@article{hendrycks2020measuring,
  title={Measuring massive multitask language understanding},
  author={Hendrycks, Dan and Burns, Collin and Basart, Steven and Zou, Andy and Mazeika, Mantas and Song, Dawn and Steinhardt, Jacob},
  journal={arXiv preprint arXiv:2009.03300},
  year={2020}
}

@article{clark2018think,
  title={Think you have solved question answering? try arc, the ai2 reasoning challenge},
  author={Clark, Peter and Cowhey, Isaac and Etzioni, Oren and Khot, Tushar and Sabharwal, Ashish and Schoenick, Carissa and Tafjord, Oyvind},
  journal={arXiv preprint arXiv:1803.05457},
  year={2018}
}

@article{guo2025deepseek,
  title={Deepseek-r1: Incentivizing reasoning capability in llms via reinforcement learning},
  author={Guo, Daya and Yang, Dejian and Zhang, Haowei and Song, Junxiao and Zhang, Ruoyu and Xu, Runxin and Zhu, Qihao and Ma, Shirong and Wang, Peiyi and Bi, Xiao and others},
  journal={arXiv preprint arXiv:2501.12948},
  year={2025}
}

@article{sun2023think,
  title={Think-on-graph: Deep and responsible reasoning of large language model on knowledge graph},
  author={Sun, Jiashuo and Xu, Chengjin and Tang, Lumingyuan and Wang, Saizhuo and Lin, Chen and Gong, Yeyun and Ni, Lionel M and Shum, Heung-Yeung and Guo, Jian},
  journal={arXiv preprint arXiv:2307.07697},
  year={2023}
}

@article{jiang2023structgpt,
  title={Structgpt: A general framework for large language model to reason over structured data},
  author={Jiang, Jinhao and Zhou, Kun and Dong, Zican and Ye, Keming and Zhao, Wayne Xin and Wen, Ji-Rong},
  journal={arXiv preprint arXiv:2305.09645},
  year={2023}
}

@inproceedings{tan2025paths,
  title={Paths-over-graph: Knowledge graph empowered large language model reasoning},
  author={Tan, Xingyu and Wang, Xiaoyang and Liu, Qing and Xu, Xiwei and Yuan, Xin and Zhang, Wenjie},
  booktitle={Proceedings of the ACM on Web Conference 2025},
  pages={3505--3522},
  year={2025}
}

@inproceedings{ma2025debate,
  title={Debate on graph: a flexible and reliable reasoning framework for large language models},
  author={Ma, Jie and Gao, Zhitao and Chai, Qi and Sun, Wangchun and Wang, Pinghui and Pei, Hongbin and Tao, Jing and Song, Lingyun and Liu, Jun and Zhang, Chen and others},
  booktitle={Proceedings of the AAAI Conference on Artificial Intelligence},
  volume={39},
  number={23},
  pages={24768--24776},
  year={2025}
}

@inproceedings{liu2025symagent,
  title={Symagent: A neural-symbolic self-learning agent framework for complex reasoning over knowledge graphs},
  author={Liu, Ben and Zhang, Jihai and Lin, Fangquan and Yang, Cheng and Peng, Min and Yin, Wotao},
  booktitle={Proceedings of the ACM on Web Conference 2025},
  pages={98--108},
  year={2025}
}

@article{li2023few,
  title={Few-shot in-context learning for knowledge base question answering},
  author={Li, Tianle and Ma, Xueguang and Zhuang, Alex and Gu, Yu and Su, Yu and Chen, Wenhu},
  journal={arXiv preprint arXiv:2305.01750},
  year={2023}
}

@article{luo2023chatkbqa,
  title={Chatkbqa: A generate-then-retrieve framework for knowledge base question answering with fine-tuned large language models},
  author={Luo, Haoran and Tang, Zichen and Peng, Shiyao and Guo, Yikai and Zhang, Wentai and Ma, Chenghao and Dong, Guanting and Song, Meina and Lin, Wei and Zhu, Yifan and others},
  journal={arXiv preprint arXiv:2310.08975},
  year={2023}
}

@inproceedings{feng2025rgr,
  title={Rgr-kbqa: Generating logical forms for question answering using knowledge-graph-enhanced large language model},
  author={Feng, Tengfei and He, Liang},
  booktitle={Proceedings of the 31st International Conference on Computational Linguistics},
  pages={3057--3070},
  year={2025}
}

@article{luo2023reasoning,
  title={Reasoning on graphs: Faithful and interpretable large language model reasoning},
  author={Luo, Linhao and Li, Yuan-Fang and Haffari, Gholamreza and Pan, Shirui},
  journal={arXiv preprint arXiv:2310.01061},
  year={2023}
}

@article{achiam2023gpt,
  title={Gpt-4 technical report},
  author={Achiam, Josh and Adler, Steven and Agarwal, Sandhini and Ahmad, Lama and Akkaya, Ilge and Aleman, Florencia Leoni and Almeida, Diogo and Altenschmidt, Janko and Altman, Sam and Anadkat, Shyamal and others},
  journal={arXiv preprint arXiv:2303.08774},
  year={2023}
}

@inproceedings{atif2023beamqa,
  title={Beamqa: Multi-hop knowledge graph question answering with sequence-to-sequence prediction and beam search},
  author={Atif, Farah and El Khatib, Ola and Difallah, Djellel},
  booktitle={Proceedings of the 46th International ACM SIGIR Conference on Research and Development in Information Retrieval},
  pages={781--790},
  year={2023}
}

@article{hu2017answering,
  title={Answering natural language questions by subgraph matching over knowledge graphs},
  author={Hu, Sen and Zou, Lei and Yu, Jeffrey Xu and Wang, Haixun and Zhao, Dongyan},
  journal={IEEE Transactions on Knowledge and Data Engineering},
  volume={30},
  number={5},
  pages={824--837},
  year={2017},
  publisher={IEEE}
}

@inproceedings{lan2020query,
  title={Query graph generation for answering multi-hop complex questions from knowledge bases},
  author={Lan, Yunshi and Jiang, Jing},
  year={2020},
  organization={Association for Computational Linguistics}
}

@inproceedings{zhang2025rule,
  title={Rule-KBQA: rule-guided reasoning for complex knowledge base question answering with large language models},
  author={Zhang, Zhiqiang and Wen, Liqiang and Zhao, Wen},
  booktitle={Proceedings of the 31st International Conference on Computational Linguistics},
  pages={8399--8417},
  year={2025}
}

@article{xu2024generate,
  title={Generate-on-graph: Treat llm as both agent and kg in incomplete knowledge graph question answering},
  author={Xu, Yao and He, Shizhu and Chen, Jiabei and Wang, Zihao and Song, Yangqiu and Tong, Hanghang and Liu, Guang and Liu, Kang and Zhao, Jun},
  journal={arXiv preprint arXiv:2404.14741},
  year={2024}
}

@article{jiang2024kg,
  title={Kg-agent: An efficient autonomous agent framework for complex reasoning over knowledge graph},
  author={Jiang, Jinhao and Zhou, Kun and Zhao, Wayne Xin and Song, Yang and Zhu, Chen and Zhu, Hengshu and Wen, Ji-Rong},
  journal={arXiv preprint arXiv:2402.11163},
  year={2024}
}

@article{luo2025graph,
  title={Graph-r1: Towards agentic graphrag framework via end-to-end reinforcement learning},
  author={Luo, Haoran and Chen, Guanting and Lin, Qika and Guo, Yikai and Xu, Fangzhi and Kuang, Zemin and Song, Meina and Wu, Xiaobao and Zhu, Yifan and Tuan, Luu Anh and others},
  journal={arXiv preprint arXiv:2507.21892},
  year={2025}
}

@article{hao2025dynasearcher,
  title={DynaSearcher: Dynamic Knowledge Graph Augmented Search Agent via Multi-Reward Reinforcement Learning},
  author={Hao, Chuzhan and Feng, Wenfeng and Zhang, Yuewei and Wang, Hao},
  journal={arXiv preprint arXiv:2507.17365},
  year={2025}
}

@article{khorashadizadeh2024research,
  title={Research trends for the interplay between large language models and knowledge graphs},
  author={Khorashadizadeh, Hanieh and Amara, Fatima Zahra and Ezzabady, Morteza and Ieng, Fr{\'e}d{\'e}ric and Tiwari, Sanju and Mihindukulasooriya, Nandana and Groppe, Jinghua and Sahri, Soror and Benamara, Farah and Groppe, Sven},
  journal={arXiv preprint arXiv:2406.08223},
  year={2024}
}

@inproceedings{zhao2024breaking,
  title={Breaking the barrier: utilizing large language models for industrial recommendation systems through an inferential knowledge graph},
  author={Zhao, Qian and Qian, Hao and Liu, Ziqi and Zhang, Gong-Duo and Gu, Lihong},
  booktitle={Proceedings of the 33rd ACM International Conference on Information and Knowledge Management},
  pages={5086--5093},
  year={2024}
}

@article{zeng2024large,
  title={Large language models for social networks: Applications, challenges, and solutions},
  author={Zeng, Jingying and Huang, Richard and Malik, Waleed and Yin, Langxuan and Babic, Bojan and Shacham, Danny and Yan, Xiao and Yang, Jaewon and He, Qi},
  journal={arXiv preprint arXiv:2401.02575},
  year={2024}
}

@article{talmor2018web,
  title={The web as a knowledge-base for answering complex questions},
  author={Talmor, Alon and Berant, Jonathan},
  journal={arXiv preprint arXiv:1803.06643},
  year={2018}
}

@inproceedings{yih2016value,
  title={The value of semantic parse labeling for knowledge base question answering},
  author={Yih, Wen-tau and Richardson, Matthew and Meek, Christopher and Chang, Ming-Wei and Suh, Jina},
  booktitle={Proceedings of the 54th Annual Meeting of the Association for Computational Linguistics (Volume 2: Short Papers)},
  pages={201--206},
  year={2016}
}

@inproceedings{gu2021beyond,
  title={Beyond iid: three levels of generalization for question answering on knowledge bases},
  author={Gu, Yu and Kase, Sue and Vanni, Michelle and Sadler, Brian and Liang, Percy and Yan, Xifeng and Su, Yu},
  booktitle={Proceedings of the web conference 2021},
  pages={3477--3488},
  year={2021}
}

@article{petrochuk1804simplequestions,
  title={Simplequestions nearly solved: A new upperbound and baseline approach. arXiv 2018},
  author={Petrochuk, M and Zettlemoyer, L},
  journal={arXiv preprint arXiv:1804.08798}
}

@book{descartes1901discourse,
  title={Discourse on method},
  author={Descartes, Ren{\'e}},
  year={1901},
  publisher={Aladdin Book Company}
}

@article{shao2024deepseekmath,
  title={Deepseekmath: Pushing the limits of mathematical reasoning in open language models},
  author={Shao, Zhihong and Wang, Peiyi and Zhu, Qihao and Xu, Runxin and Song, Junxiao and Bi, Xiao and Zhang, Haowei and Zhang, Mingchuan and Li, YK and Wu, Yang and others},
  journal={arXiv preprint arXiv:2402.03300},
  year={2024}
}

@article{li2023chain,
  title={Chain of knowledge: A framework for grounding large language models with structured knowledge bases},
  author={Li, Xingxuan and Zhao, Ruochen and Chia, Yew Ken and Ding, Bosheng and Bing, Lidong and Joty, Shafiq and Poria, Soujanya},
  journal={arXiv preprint arXiv:2305.13269},
  volume={3},
  year={2023}
}

@article{baek2023knowledge,
  title={Knowledge-augmented language model prompting for zero-shot knowledge graph question answering},
  author={Baek, Jinheon and Aji, Alham Fikri and Saffari, Amir},
  journal={arXiv preprint arXiv:2306.04136},
  year={2023}
}

@inproceedings{bollacker2008freebase,
  title={Freebase: a collaboratively created graph database for structuring human knowledge},
  author={Bollacker, Kurt and Evans, Colin and Paritosh, Praveen and Sturge, Tim and Taylor, Jamie},
  booktitle={Proceedings of the 2008 ACM SIGMOD international conference on Management of data},
  pages={1247--1250},
  year={2008}
}

@article{shi2025search,
  title={Search and Refine During Think: Autonomous Retrieval-Augmented Reasoning of LLMs},
  author={Shi, Yaorui and Li, Sihang and Wu, Chang and Liu, Zhiyuan and Fang, Junfeng and Cai, Hengxing and Zhang, An and Wang, Xiang},
  journal={arXiv preprint arXiv:2505.11277},
  year={2025}
}

@article{team2024qwen2,
  title={Qwen2 technical report},
  author={Team, Qwen and others},
  journal={arXiv preprint arXiv:2407.10671},
  volume={2},
  pages={3},
  year={2024}
}

@article{xue2024decompose,
  title={Decompose, analyze and rethink: Solving intricate problems with human-like reasoning cycle},
  author={Xue, Shangzi and Huang, Zhenya and Liu, Jiayu and Lin, Xin and Ning, Yuting and Jin, Binbin and Li, Xin and Liu, Qi},
  journal={Advances in Neural Information Processing Systems},
  volume={37},
  pages={357--385},
  year={2024}
}

@article{schick2023toolformer,
  title={Toolformer: Language models can teach themselves to use tools},
  author={Schick, Timo and Dwivedi-Yu, Jane and Dess{\`\i}, Roberto and Raileanu, Roberta and Lomeli, Maria and Hambro, Eric and Zettlemoyer, Luke and Cancedda, Nicola and Scialom, Thomas},
  journal={Advances in Neural Information Processing Systems},
  volume={36},
  pages={68539--68551},
  year={2023}
}

@article{raffel2020exploring,
  title={Exploring the limits of transfer learning with a unified text-to-text transformer},
  author={Raffel, Colin and Shazeer, Noam and Roberts, Adam and Lee, Katherine and Narang, Sharan and Matena, Michael and Zhou, Yanqi and Li, Wei and Liu, Peter J},
  journal={Journal of machine learning research},
  volume={21},
  number={140},
  pages={1--67},
  year={2020}
}

@article{jiang2022unikgqa,
  title={Unikgqa: Unified retrieval and reasoning for solving multi-hop question answering over knowledge graph},
  author={Jiang, Jinhao and Zhou, Kun and Zhao, Wayne Xin and Wen, Ji-Rong},
  journal={arXiv preprint arXiv:2212.00959},
  year={2022}
}

@inproceedings{xiong2017explicit,
  title={Explicit semantic ranking for academic search via knowledge graph embedding},
  author={Xiong, Chenyan and Power, Russell and Callan, Jamie},
  booktitle={Proceedings of the 26th international conference on world wide web},
  pages={1271--1279},
  year={2017}
}

@inproceedings{luo2025chatrule,
  title={Chatrule: Mining logical rules with large language models for knowledge graph reasoning},
  author={Luo, Linhao and Ju, Jiaxin and Xiong, Bo and Li, Yuan-Fang and Haffari, Gholamreza and Pan, Shirui},
  booktitle={Pacific-Asia Conference on Knowledge Discovery and Data Mining},
  pages={314--325},
  year={2025},
  organization={Springer}
}

@article{wei2022chain,
  title={Chain-of-thought prompting elicits reasoning in large language models},
  author={Wei, Jason and Wang, Xuezhi and Schuurmans, Dale and Bosma, Maarten and Xia, Fei and Chi, Ed and Le, Quoc V and Zhou, Denny and others},
  journal={Advances in neural information processing systems},
  volume={35},
  pages={24824--24837},
  year={2022}
}

@article{liu2025visual,
  title={Visual-rft: Visual reinforcement fine-tuning},
  author={Liu, Ziyu and Sun, Zeyi and Zang, Yuhang and Dong, Xiaoyi and Cao, Yuhang and Duan, Haodong and Lin, Dahua and Wang, Jiaqi},
  journal={arXiv preprint arXiv:2503.01785},
  year={2025}
}

@article{jin2025search,
  title={Search-r1: Training llms to reason and leverage search engines with reinforcement learning},
  author={Jin, Bowen and Zeng, Hansi and Yue, Zhenrui and Yoon, Jinsung and Arik, Sercan and Wang, Dong and Zamani, Hamed and Han, Jiawei},
  journal={arXiv preprint arXiv:2503.09516},
  year={2025}
}

@article{xie2023pixiu,
  title={Pixiu: A comprehensive benchmark, instruction dataset and large language model for finance},
  author={Xie, Qianqian and Han, Weiguang and Zhang, Xiao and Lai, Yanzhao and Peng, Min and Lopez-Lira, Alejandro and Huang, Jimin},
  journal={Advances in Neural Information Processing Systems},
  volume={36},
  pages={33469--33484},
  year={2023}
}

@article{xie2024finben,
  title={Finben: A holistic financial benchmark for large language models},
  author={Xie, Qianqian and Han, Weiguang and Chen, Zhengyu and Xiang, Ruoyu and Zhang, Xiao and He, Yueru and Xiao, Mengxi and Li, Dong and Dai, Yongfu and Feng, Duanyu and others},
  journal={Advances in Neural Information Processing Systems},
  volume={37},
  pages={95716--95743},
  year={2024}
}

@article{xie2025medical,
  title={Medical foundation large language models for comprehensive text analysis and beyond},
  author={Xie, Qianqian and Chen, Qingyu and Chen, Aokun and Peng, Cheng and Hu, Yan and Lin, Fongci and Peng, Xueqing and Huang, Jimin and Zhang, Jeffrey and Keloth, Vipina and others},
  journal={npj Digital Medicine},
  volume={8},
  number={1},
  pages={141},
  year={2025},
  publisher={Nature Publishing Group UK London}
}

@article{wang2023pre,
  title={Pre-trained language models in biomedical domain: A systematic survey},
  author={Wang, Benyou and Xie, Qianqian and Pei, Jiahuan and Chen, Zhihong and Tiwari, Prayag and Li, Zhao and Fu, Jie},
  journal={ACM Computing Surveys},
  volume={56},
  number={3},
  pages={1--52},
  year={2023},
  publisher={ACM New York, NY}
}

@inproceedings{tan2025reason,
  title={Reason-rft: Reinforcement fine-tuning for visual reasoning of vision language models},
  author={Tan, Huajie and Ji, Yuheng and Hao, Xiaoshuai and Chen, Xiansheng and Wang, Pengwei and Wang, Zhongyuan and Zhang, Shanghang},
  booktitle={The Thirty-ninth Annual Conference on Neural Information Processing Systems},
  year={2025}
}

\appendix

\section{APPENDIX}
\subsection{SFT Dataset Construction}
\label{app:data_construct}
We construct a Long CoT dataset comprising 4862 samples by prompting QwQ-32B with KG and Web retrieval tools. For quality filtering, we formulate two stages filtering processes:\\
\noindent $\blacktriangleright$\textbf{Format Check.} COT trajectories must not contain any labels except the defined ones, and the <plan></plan> can only be invoked once.\\
\noindent $\blacktriangleright$\textbf{Correctness Check.}
\begin{itemize}[itemsep=0.5ex, leftmargin=3mm]
\item \textbf{Answer Correctness}: The answer contained within the label must be correct; otherwise, the trajectory will be filtered out.
\item \textbf{Retrieval Correctness}: When the knowledge graph is complete: The COT must not contain the <web\_search>, and the <neighbor\_information> must include the correct answer. If not, the trajectory will be filtered out. When the knowledge graph is incomplete: The COT contains the <web\_search> and the <web\_information>  must includes the correct answer. If not, the trajectory will be filtered out.
\item \textbf{Plan Correctness}: GPT-4 is used to determine whether the decomposition and planning in the <plan> section are reasonable. A score of 1 is assigned for reasonable planning, and 0 for unreasonable planning. Trajectories with a score of 0 are filtered out.
\end{itemize}

\subsection{Baselines}
\label{app:baseline}
\begin{sloppypar}
We assess the performance of Graph-RFT, leveraging Qwen2.5-7B as the backbone model. The baselines we compare can be divided into three groups: (1) LLM-only methods, including standard In-Context (IO) prompting, Chain-of-Thought (CoT) prompting, and Self-Consistency (SC) prompting. All comparisons utilize six in-content examples and incorporate "step-by-step" reasoning chains. (2) Semantic Parsing methods, including KB-BINDER\cite{li2023few}, ChatKBQA\cite{luo2023chatkbqa}. (3) Retrieval Augmented methods, including StructGPT\cite{jiang2023structgpt}, RoG\cite{luo2023reasoning}, ToG\cite{sun2023think}, PoG\cite{tan2025paths}, GoG\cite{xu2024generate}. 
\end{sloppypar}

\subsection{Implementation Details}
\label{app:imp}
We use Qwen2.5-7B-instruct and Qwen2.5-7B-base as backbones \cite{team2024qwen2} to obtain Graph-RFT-instruct and Graph-RFT-base, respectively, using PyTorch on 8 NVIDIA A800 (80GB) GPUs. In the SFT stage, we leverage LLaMA-Factory framework for efficient LLM fine-tuning. We employ a batchsize of 256 for 2 epochs with a learning rate of 1.4e-5 and AdamW optimizer with cosine decay. In the RL stage, we adopt the Verl framework with the batch size of 128 while the mini-batch size of 32, a rollout number of 8.
\subsection{Prompt for Graph-RFT}
\label{app:prompt}
\begin{figure}[H]
    \centering
    \includegraphics[width=.8\linewidth]{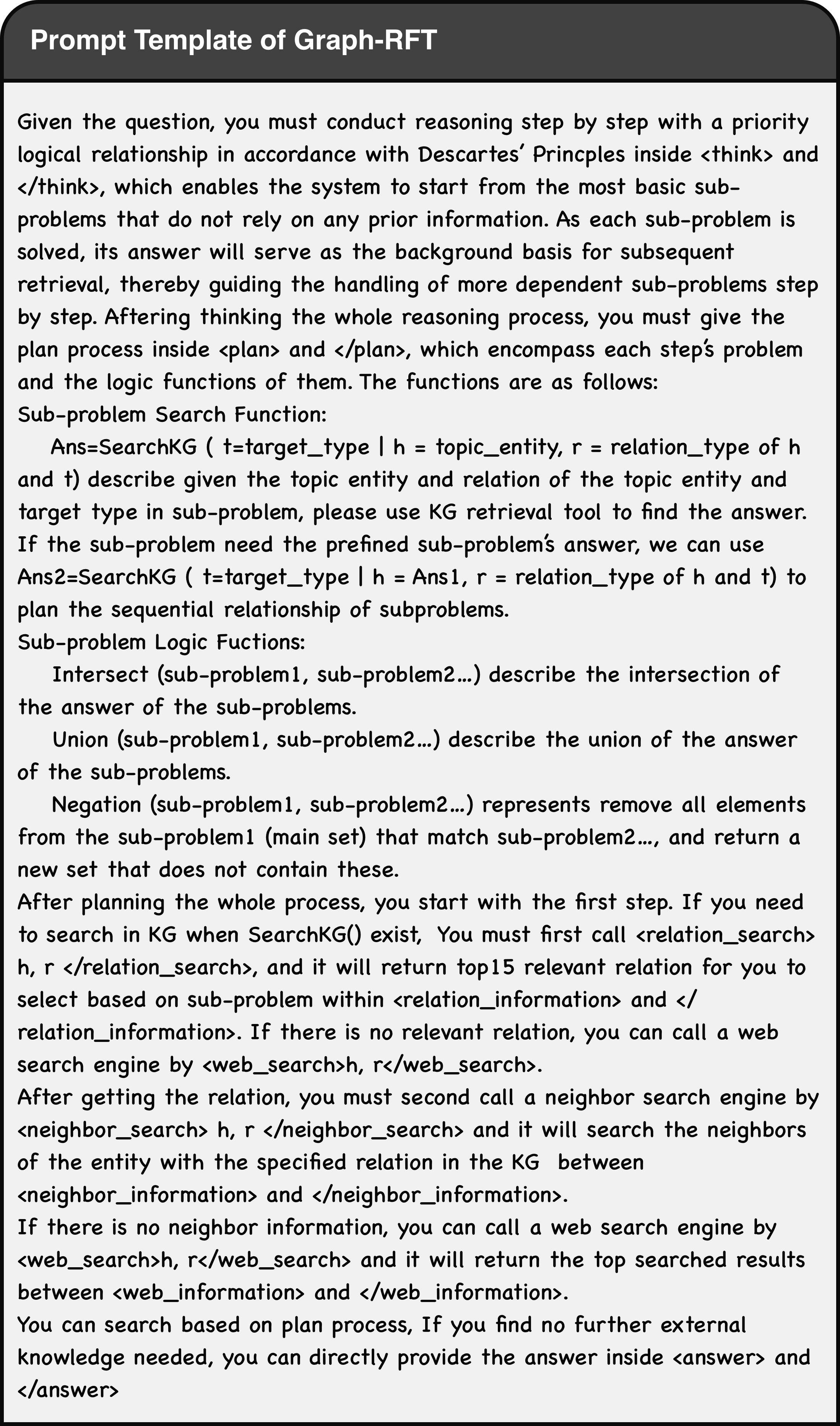}
    \caption{Prompt template of our Graph-RFT.}
    \label{fig:prompt}
\end{figure}

In this section, we present the prompt template. The prompt comprises core components that guide the reasoning and interaction process with the knowledge graph. As illustrated in Figure \ref{fig:prompt}, we adopt a structured interaction format, which utilizes the <plan>, <relation\_search>, <neighbor\_search>, <web\_search>, and <answer>.
To facilitate complex reasoning, Graph-RFT first formulates a global plan. This plan outlines how to decompose the original question into manageable sub-problems using the custom-designed logic functions. Following the planning phase, each sub-problem is retrieved step-by-step from the KGs, with the invocation of dedicated KG retrieval tools.
Subsequently, Graph-RFT employs a clearly defined set of tools to interact with both the knowledge graph and external documents. Through this interaction, it performs multi-step reasoning and ultimately derives the final answer. By integrating structured knowledge from the incomplete KG with supplementary information from external sources, this systematic approach enables Graph-RFT to effectively address complex problems.

\subsection{Different Knowledge Bases for Graph-RFT}
\label{knowledege_bases}
As shown in Table \ref{app:knowledge_base}, Graph-RFT achieves significant improvements with different source KGs on CWQ and WebQSP in IKG-40\%, compared to ToG. On the other hand, different source KGs might have different effects on the performance of ToG. Notably, Freebase brings more significant improvements on CWQ and WebQSP than Wikidata, since both datasets are constructed upon Freebase. 

\begin{table}[htbp]
  \centering
  \caption{Performances of Graph-RFT using different source KGs on CWQ and WebQSP in IKG-40\%.}
  \label{app:knowledge_base}
  \resizebox{0.8\linewidth}{!}{
    \begin{tabular}{c|cc}
      \toprule
      \textbf{Method} &\textbf{CWQ} & \textbf{WebQSP}   \\
      \midrule
      w/Freebase (ToG)&34.9 & 58.4  \\
      w/WikiData (ToG)&32.6&54.7\\\midrule
      w/Freebase (Graph-RFT) & 67.2 & 86.3 \\
      w/WikiData (Graph-RFT)& 61.5 & 77.8\\
      \bottomrule
    \end{tabular}
  }
\end{table}

\subsection{Case Study}
\label{case_study}
\begin{figure}[H]
    \centering
    \includegraphics[width=.9\linewidth]{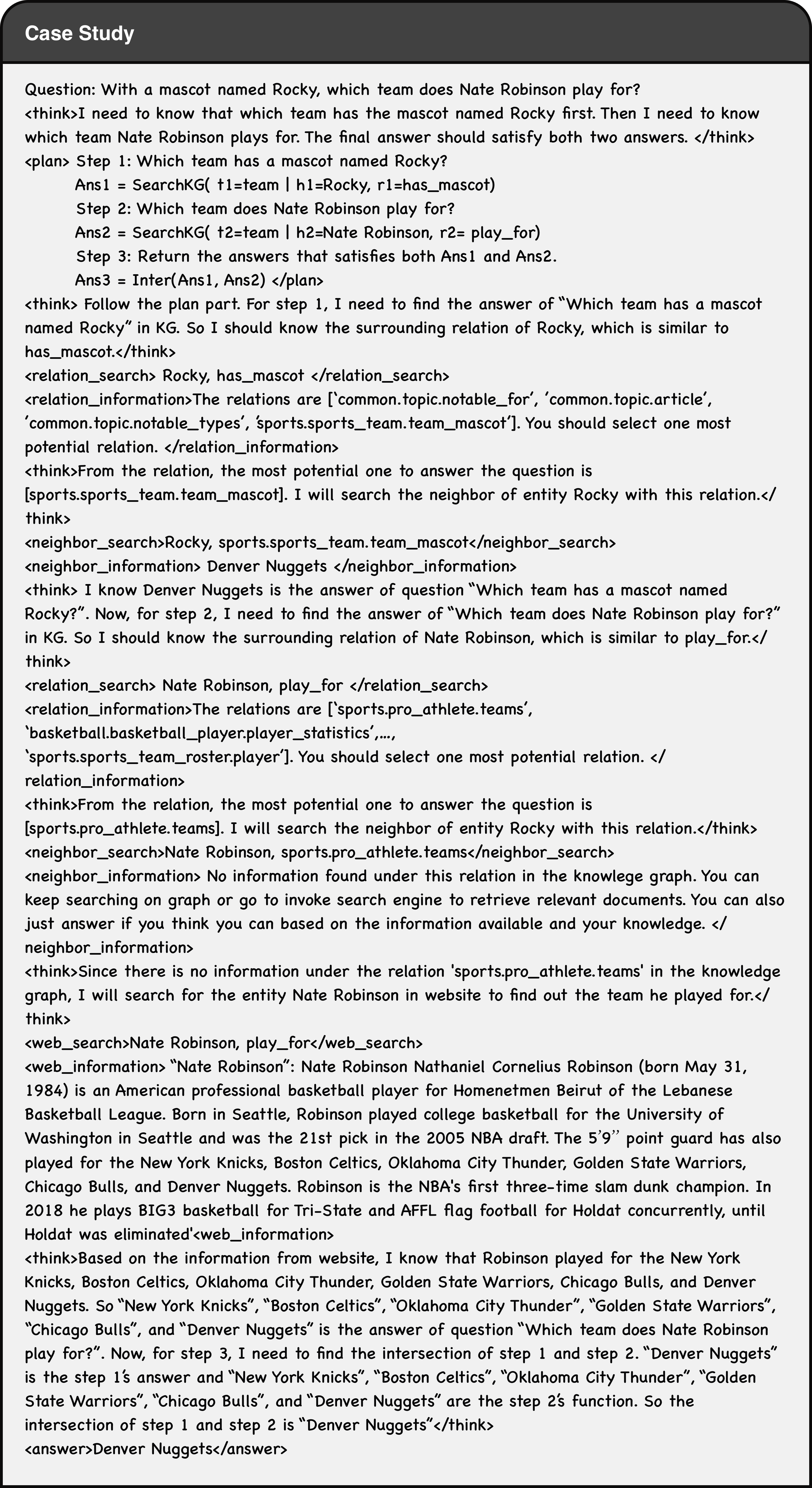}
    \caption{Case Study.}
    \label{fig:case}
\end{figure}

To demonstrate Graph-RFT’s real-world operation, we provide a detailed case study that breaks down its reasoning process when addressing complex questions. As shown in Figure \ref{fig:case}, this example focuses on a team-related problem, illustrating how Graph-RFT systematically processes the question by leveraging both the knowledge graph and web resources.

\begin{figure*}[t!]
    \centering
    \includegraphics[width=.9\textwidth]{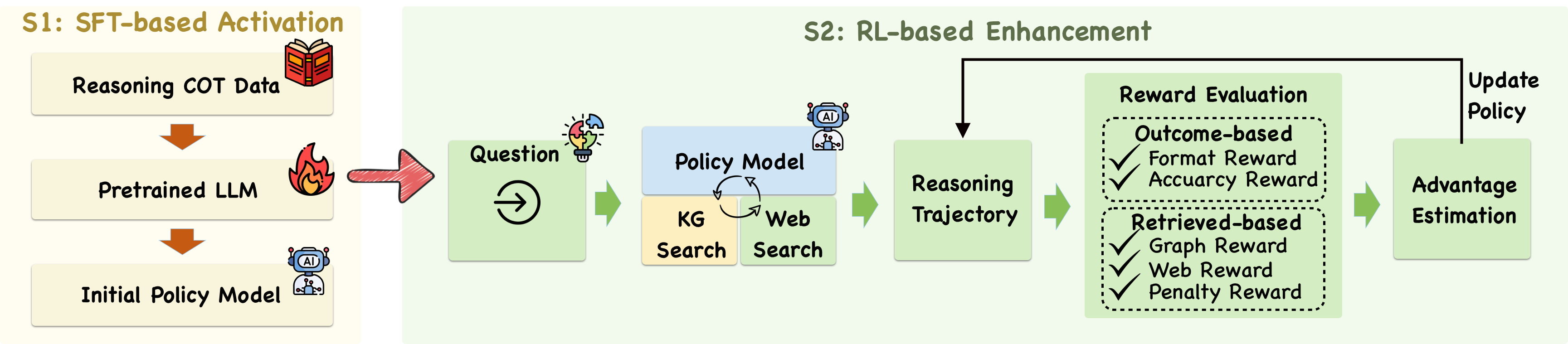}
    \caption{Graph-RFT Training Framework: SFT-based Reasoning Activation (S1) and RL-based Reasoning Enhancement (S2).}
    \label{frame}
    \vspace{-2ex}
\end{figure*}

\begin{figure*}[t!]
    \centering
    \includegraphics[width=.8\textwidth]{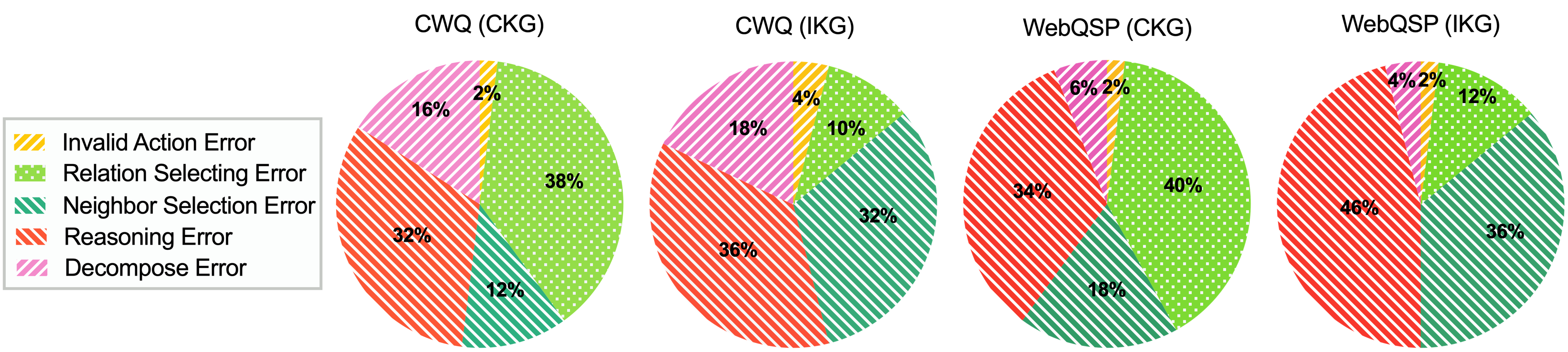}
    \caption{Error Type}
    \label{fig:error}
    \vspace{-2ex}
\end{figure*}

\end{document}